\documentclass{cambrian}
\usepackage[utf8]{inputenc} 
\usepackage[T1]{fontenc}    
\usepackage{hyperref}       
\usepackage{url}            
\usepackage{booktabs}       
\usepackage{amsfonts}       
\usepackage{nicefrac}       
\usepackage{microtype}      

\usepackage{mwe}
\usepackage{graphicx}
\usepackage{pifont} 
\usepackage{amssymb}
\usepackage{adjustbox}
\usepackage{colortbl}
\usepackage{array}
\usepackage{multirow}
\usepackage{makecell}
\usepackage{bbm}
\usepackage{collcell,xfp}
\usepackage{pgf}

\usepackage{tikz}
\usepackage[most]{tcolorbox}
\usepackage{csquotes}
\usepackage[noorphans,vskip=1em,leftmargin=1em]{quoting}
\usepackage{pifont} 
\usepackage{enumitem} 
\usepackage{tcolorbox} 
\usepackage{forest}
\usepackage{wrapfig}
\usepackage{caption}
\usepackage{subcaption}
\usepackage{longtable}
\usepackage{algpseudocode}
\usepackage{amsmath}
\usepackage[capitalize]{cleveref}
\usepackage[ruled,vlined,linesnumbered]{algorithm2e} 
\usepackage[percent]{overpic}
\usepackage{xspace}
\usepackage[normalem]{ulem}  
\usepackage{tocloft}  
\usepackage[sort]{natbib}  

\newcommand{\cmark}{\ding{51}}
\newcommand{\xmark}{\ding{55}}

\usepackage{fontawesome}
\newcommand{\huggingface}{\raisebox{-1.5pt}{\includegraphics[height=1.05em]{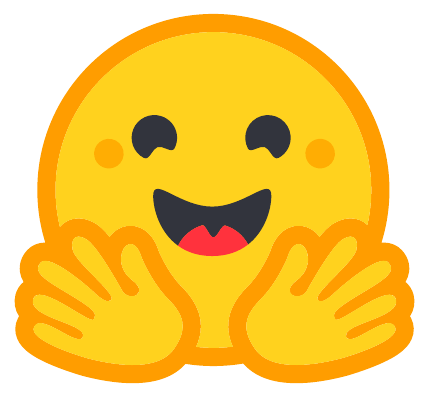}}\xspace}
\newcommand{\github}{\raisebox{-1.5pt}{\includegraphics[height=1.05em]{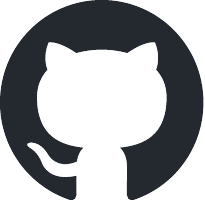}}\xspace}

\newcommand{\worldwideweb}{\raisebox{-1.5pt}{\includegraphics[height=1.05em]{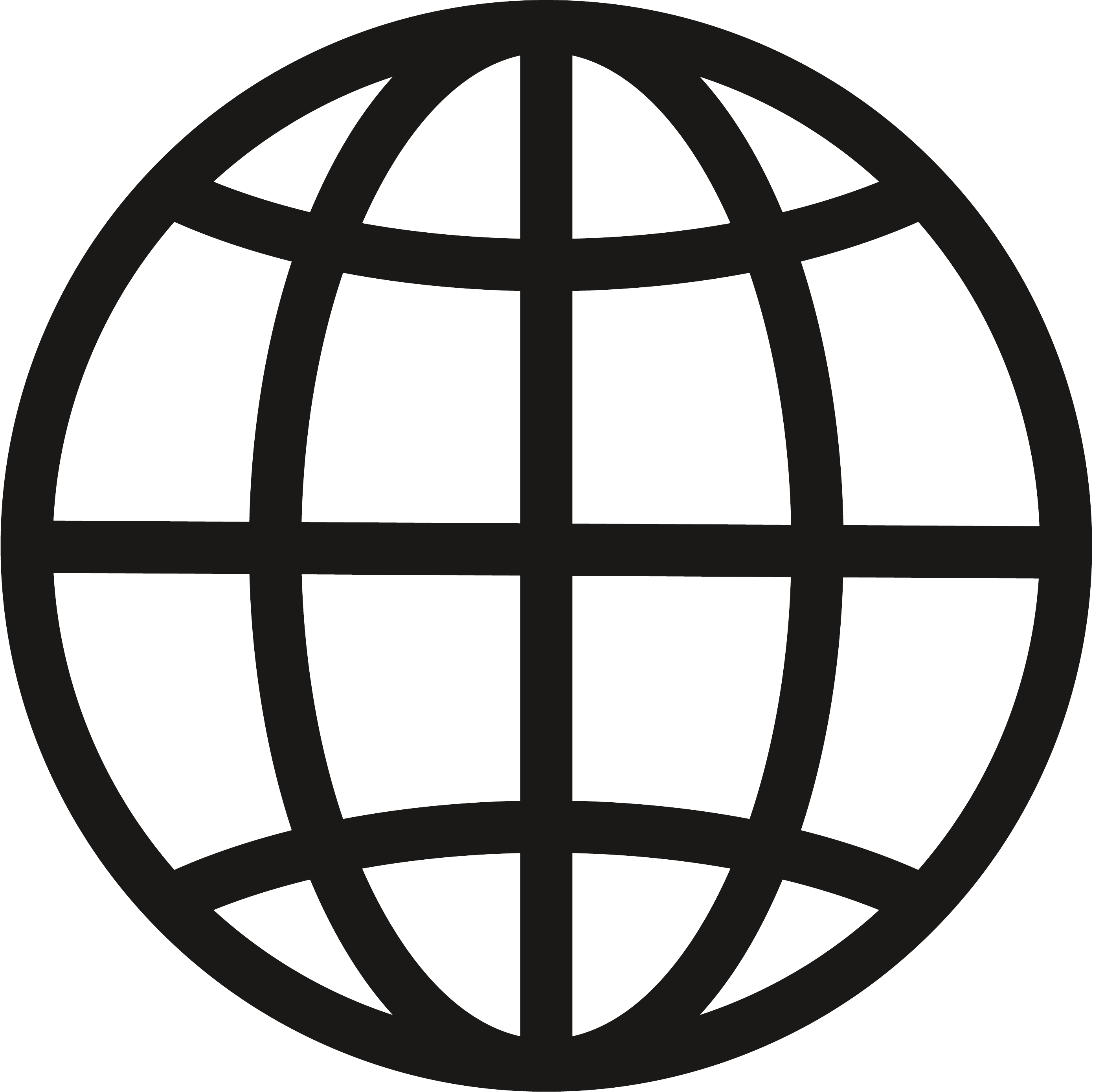}}\xspace}

\definecolor{scholarblue}{rgb}{0.21,0.49,0.74}
\definecolor{bluelink}{RGB}{0,113,188}
\definecolor{greenlink}{RGB}{0,188,113}
\hypersetup{
    colorlinks=true,%
    citecolor=scholarblue,%
    filecolor=red,%
    linkcolor=red!93!black,%
    urlcolor=bluelink
}

\usepackage{titlesec}
\titlespacing*{\paragraph}
    {0pt}     
    {0.25em}  
    {1em}  

\definecolor{navyblue}{HTML}{0071BC}
\iftrue   
\newcommand{\displaytodo}[1]{#1}
\else
\newcommand{\displaytodo}[1]{}
\fi

\newcommand{\model}{Cambrian-\emph{P}\xspace}
\newcommand{\cambrians}{Cambrian-\emph{S}\xspace}
\newcommand{\vsidata}{VSI-590K\xspace}

\def\eg{\emph{e.g.}} 
\def\ie{\emph{i.e.}}

\definecolor{blindcolor}{HTML}{AB2AC6}    
\definecolor{chancecolor}{HTML}{F59E0B}   
\definecolor{singlecolor}{HTML}{06B6D4}   
\definecolor{multiplecolor}{HTML}{2563EB} 
\definecolor{captioncolor}{HTML}{22C55E}  

 \newcommand{\culine}[2]{%
    \def\temp@uline{\bgroup\markoverwith
        {\textcolor{#1}{\rule[-0.5ex]{2pt}{1pt}}}\ULon}%
    \temp@uline{#2}%
}
 \newcommand{\cthickuline}[3][0.8pt]{%
    \def\temp@uline{\bgroup\markoverwith
        {\textcolor{#2}{\rule[-0.5ex]{2pt}{#1}}}\ULon}%
    \temp@uline{#3}%
}

\title{\center{Cambrian-\emph{P}: Pose-Grounded Video Understanding}}

\renewcommand\Affilfont{\normalfont\fontsize{11}{15}\selectfont\centering}
\author{
    Jihan~Yang\textsuperscript{1}$^{*}$ \quad\ 
    Zifan~Zhao\textsuperscript{1}$^{*}$ \quad\ 
    Xichen~Pan\textsuperscript{1} \quad\
    Shusheng~Yang\textsuperscript{1} \quad 
    Junyi~Zhang\textsuperscript{2} \quad\ \quad \quad
    Bingyi~Kang\textsuperscript{1} \quad
    Hu~Xu\textsuperscript{3} \quad
    Shang-Wen~Li\textsuperscript{3} \quad
    Saining~Xie\textsuperscript{1} \\
    \textsuperscript{1} New York University \quad \textsuperscript{2} UC Berkeley \quad \textsuperscript{3} Meta FAIR
}

\begin{abstract}
Camera pose matters. The position and orientation of each viewpoint define a shared spatial coordinate frame that relates observations across video frames. Yet this signal is largely absent from multimodal LLMs (MLLMs) for video understanding, 
which process frames as isolated 2D snapshots, instead of the persistent scene humans perceive.
We revisit pose as a lightweight supervisory signal and introduce \model, a video MLLM augmented with per-frame camera pose tokens instantiated from two learnable queries and a pose regression head. With a carefully designed sampling scheme, the model achieves substantial 4.5–6.5 point gains on spatial reasoning benchmarks such as VSI-Bench, generalizes across eight additional spatial and general video QA benchmarks, and, as a byproduct, achieves the best ScanNet ATE among streaming pose estimation methods. 
Surprisingly, training on pseudo-annotated poses from in-the-wild video further improves general video QA benchmarks, showing pose helps beyond spatial reasoning.
Together, these results position camera pose as a fundamental signal for video models that reason about the physical world.

\end{abstract}

\renewcommand{\thefootnote}{\arabic{footnote}}  

\begin{document}

\maketitle

\begingroup
    \renewcommand{\thefootnote}{\fnsymbol{footnote}}
    \footnotetext[1]{JY led the project, JY and ZZ contributed equally.}
\endgroup

\begin{center}
    \renewcommand{\arraystretch}{1.5}
    \begin{tabular}{rll}
        \worldwideweb{} & \textbf{Website} & \url{https://cambrian-mllm.github.io}\\
        \github{} & \textbf{Code} & \url{https://github.com/cambrian-mllm/cambrian-p}\\
        \huggingface{} & \textbf{\model Models} & \href{https://huggingface.co/collections/nyu-visionx/cambrian-p-models}{\nolinkurl{https://hf.co/collections/nyu-visionx/cambrian-p}} \\
        \huggingface{} & \textbf{Data} & \url{https://huggingface.co/datasets/nyu-visionx/Cambrian-P-Data} \\
    \end{tabular}
\end{center}


\newpage
{
    \hypersetup{linkcolor=black}
    \tableofcontents
}
\newpage


\section{Introduction}
\label{sec:intro}

Video is the projection of a dynamic 3D scene from a coherent sequence of viewpoints. Each viewpoint is defined by the observer's pose, \textit{i.e.}, its 3D position ($\mathbb{R}^3$) and orientation ($SO(3)$), specifying how the camera is embedded in the physical world~\cite{hartley2003multiple,ma2005invitation}. Pose provides the link between pixels and geometry, serving as a global anchor that relates distinct views to a shared coordinate frame~\cite{hartley2003multiple,schonberger2016structure}. With pose, a video is no longer a collection of ambiguous image projections, but a coherent 3D scene~\cite{schonberger2016structure,szeliski2022computer}. 

Recovering camera poses from visual observations has therefore been a cornerstone of 3D vision and robotics for decades. 
Structure from Motion (SfM) emerged in the 1980s precisely for this purpose~\cite{longuet1981computer,schonberger2016structure}, and remains a prerequisite for a wide array of downstream tasks: multi-view stereo~\cite{seitz2006comparison}, 3D reconstruction, and neural rendering with NeRF~\cite{mildenhall2021nerf} or 3D Gaussian Splatting~\cite{kerbl20233d}. In parallel, robotics and augmented-reality systems depend on SLAM~\cite{cadena2017past} and visual odometry~\cite{nister2004visual} to localize themselves and reason about their own motion. More recently, the community has recognized that jointly predicting pose and depth with feed-forward transformers can recover dense 3D structure in a single pass~\cite{wang2025vggt,lin2025depth}, further underscoring the centrality of pose as a geometric primitive.

Yet, the role of pose has remained confined to 3D vision. We argue that its value extends far beyond. Multimodal large language models (MLLMs)~\cite{brown2020language,touvron2023llama,touvron2023llama2,bai2023qwen,grattafiori2024llama,achiam2023gpt,liu2023visual,tong2024cambrian} now excel at semantic video understanding—recognizing actions, summarizing narratives, and answering questions—but consistently struggle when tasks demand spatial reasoning~\cite{team2025gemini,kim2024openvla,yang2024virl,yang2024think}. We contend that this failure is not incidental. Without explicit grounding in 3D geometry, each frame is processed as an independent 2D snapshot, disconnected from the spatial structure across views. Our key insight is that pose naturally closes this gap. It is the lightest 3D signal, compactly encoding how views relate geometrically; it enforces global consistency through rigid-body (SE(3)) constraints; and it disentangles camera motion from scene dynamics, collapsing the space of plausible spatial interpretations.
{
These properties also underpin human vision: viewers naturally separate their own motion from motion in the scene, and maintain a coherent 3D world across viewpoints.
Together, they make pose not merely a useful auxiliary cue but a foundational inductive bias for video understanding.
}

\begin{figure}[t]
  \centering
  \includegraphics[width=1\textwidth]{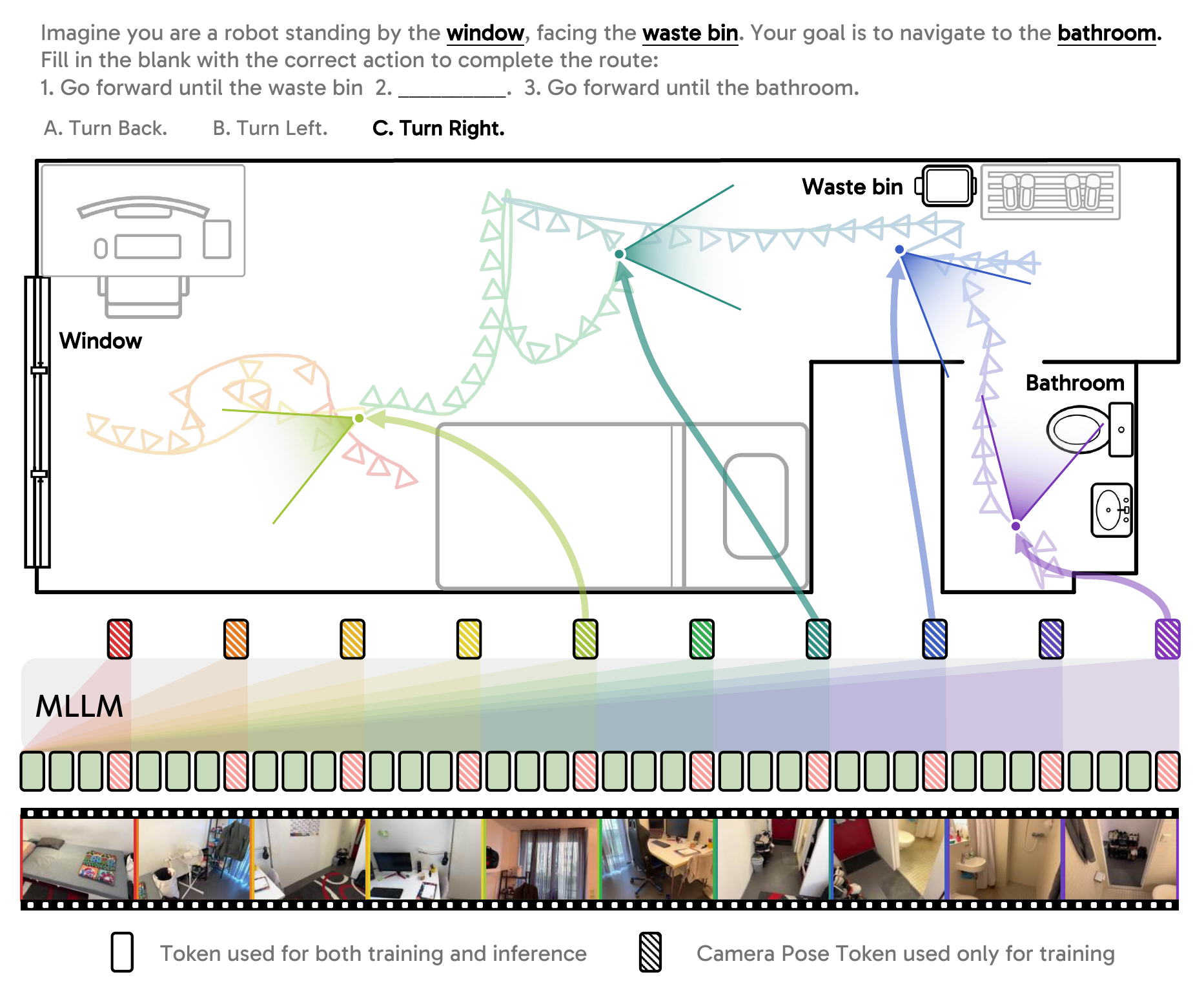}
  \caption{\textbf{\model illustration in video QA}. \model equips the current video understanding paradigm with native camera pose prediction using per-frame camera pose tokens instantiated from two learnable queries. \model positions video frames into a shared spatial coordinate frame, then effectively models the underlying 3D world projected in video. For VQA, the pose tokens are used during training and omitted by default at inference.}
  \label{fig:teaser}
\end{figure}

In this work, we present \model, which grounds pixels with camera poses by introducing pose supervision into MLLMs, calling for a new paradigm for video understanding. \model introduces minimal architectural overhead: one camera pose token, instantiated from one of two learnable queries, is appended to each frame's visual features, and a lightweight projector and head regress pose parameters from the LLM's hidden states. However, naively adding a pose regression loss does not work out of the box. Through careful investigation, we find that pose estimation and video question answering rely on fundamentally different frame sampling and data augmentation strategies. Uniform temporal sampling common in video understanding provides a shortcut for memorizing poses, while the heavy augmentation typical in pose estimation distracts from semantic comprehension. To reconcile these tensions, we design an interleaved training strategy that incorporates pose-only training data processed and augmented following its preferred practices, along with a random-jitter frame sampling strategy that introduces controlled perturbation to the conventional uniform sampling used for VQA.

With this lightweight design and optimized joint training, \model achieves state-of-the-art spatial VQA performance and competitive streaming camera pose estimation. On VSI-Bench~\cite{yang2024think} and VSTI-Bench~\cite{fan2025vlm}, \model obtains 4.5–6.5 point gains over its no-pose counterpart, and demonstrates strong out-of-distribution generalization across eight spatial and general VQA benchmarks, including MindCube~\cite{yin2025spatial}, MVBench~\cite{li2024mvbench}, and EgoSchema~\cite{mangalam2023egoschema}. Scaling pose supervision to in-the-wild videos with pseudo annotations further improves general video QA benchmarks, {positioning pose as a promising signal for video understanding beyond spatial reasoning.} For pose estimation, \model delivers the best ScanNet ATE among streaming camera pose estimation methods, outperforming specialist reconstruction models such as StreamVGGT~\cite{zhuo2026streaming}, CUT3R~\cite{wang2025continuous}, and Point3R~\cite{wu2025point3r}. We further observe consistent scaling with model size, data size, and training iterations for both tasks. Finally, our analysis reveals two additional insights: (i) while adding depth supervision may seem intuitive for injecting 3D priors, it proves suboptimal for VQA compared to learning from camera pose; and (ii) camera pose enables MLLMs to think more globally across video frames, evidenced by the significant improvement of \model when answering questions about spatially distant objects.

\section{Related Work}
\noindent\textbf{Multimodal Large Language Models}
Driven by the tremendous success of Large Language Models (LLMs)~\cite{brown2020language,touvron2023llama,touvron2023llama2,bai2023qwen,grattafiori2024llama,achiam2023gpt} in linguistic understanding and reasoning, alongside powerful pretrained visual representations~\cite{radford2021learning,he2022masked,oquab2023dinov2,zhai2023sigmoid,tschannen2025siglip}, Multimodal Large Language Models (MLLMs)~\cite{li2024llava,bai2023qwenvl,li2023blip2} extend LLMs beyond language-only corpora and have achieved impressive progress in understanding visual media such as images~\cite{liu2024improved,tong2024cambrian,li2024llava,team2023gemini,chen2024internvl,li2025eagle,chen2025eagle} and videos~\cite{yang2026towards,wang2024qwen2vl,Qwen3-VL,zhang2024video,ren2024timechat,song2024moviechat}. However, despite their remarkable success in semantic parsing~\cite{chen2015microsoft,agrawal2019nocaps}, world knowledge acquisition~\cite{yue2024mmmu,hu2025video,saikh2022scienceqa}, and general reasoning~\cite{lu2023mathvista,yue2025mmmu}, MLLMs are still far from achieving human-level embodied intelligence capable of perceiving, reasoning, and acting within the 3D real world~\cite{team2025gemini,kim2024openvla,yang2024virl}.
Recent studies~\cite{yang2024think,ramakrishnan2024does,yin2025spatial,yeh2025seeing,brown2025shortcuts} pinpoint that a fundamental deficit hindering existing MLLMs from this goal is their unsatisfactory visual spatial intelligence, which serves as one of the foundational elements for humans to understand the 3D outside world but remains largely absent in modern MLLMs.
This paper aims to bridge this gap.

\noindent\textbf{Visual Spatial Intelligence}
The growing interest in grounding MLLMs in the real 3D world has created an urgent need to improve their visual spatial intelligence: the ability to understand underlying spatial geometry from visual inputs. Motivated by this, recent works have proposed various benchmarks to evaluate this capability using single-image~\cite{ramakrishnan2024does}, multi-image~\cite{yin2025spatial}, or video inputs~\cite{yang2024think} (which is the primary focus of our work). Their results suggest that even frontier MLLMs still fall significantly behind human performance in spatial understanding.
To bridge this gap, several studies~\cite{yang2026towards,brown2025simsv,fan2025vlm,yang2025visual,chen2024spatialvlm} curate spatial-oriented data by repurposing existing 3D-related datasets~\cite{dai2017scannet,yeshwanth2023scannet++,dehghan2021arkitscenes,roberts2021hypersim,armeni20163d}, applying pseudo-labeling, or designing synthetic data generation pipelines~\cite{deitke2022ProcTHOR}. These efforts not only improve models' spatial understanding but also provide foundational datasets for future exploration.
\cite{ouyang2025spacer,yang2025visual,liu2025spatial} propose to finetune MLLMs on spatial data using reinforcement learning to improve their spatial reasoning capability.
Another line of research~\cite{fan2025vlm,li2026thinking,zheng2025learning} introduces 3D features from off-the-shelf 3D encoders~\cite{wang2025vggt}.
While this significantly improves MLLMs' spatial awareness, the approach remains inflexible as it is largely constrained by the quality of the pre-trained features. Recent work~\cite{hu2025g} unifies 3D reconstruction with spatial understanding with a dual-encoder and mixture-of-transformers design, which is heavy and yields suboptimal results.  

\noindent\textbf{Camera Pose Estimation}
Camera pose estimation serves as a pillar of 3D vision. It is not merely an isolated task, but the prerequisite for a wide spectrum of downstream applications, ranging from dense multi-view reconstruction~\cite{schonberger2016structure,furukawa2009accurate,yao2018mvsnet} to modern neural rendering~\cite{mildenhall2021nerf,kerbl20233d} and robotic navigation~\cite{qin2018vins,cadena2017past}. 
Traditionally, recovering camera extrinsics relies on SfM and SLAM systems~\cite{hartley2003multiple,schonberger2016structure,mur2015orb}. While mathematically elegant, these heuristic-based pipelines frequently struggle in ill-posed scenarios characterized by textureless regions, repetitive patterns, or dynamic environments.
Recently, a paradigm shift toward data-driven, feed-forward 3D estimation has emerged, with methods like DUSt3R~\cite{wang2024dust3r} and MASt3R~\cite{leroy2024grounding} bypassing fragile heuristics via direct dense pointmap regression, giving rise to a broad family of follow-up works. 
Among them, offline models~\cite{wang2025vggt,wang2026pi,lin2025depth,keetha2025mapanything} jointly process multiple views and typically offer stronger bidirectional reasoning over the full observation set, while streaming approaches~\cite{wang20253d,wang2025continuous,zhuo2026streaming} process frames incrementally, making them better suited for arbitrary-length videos.

In this work, we contextualize the data-driven learning of 3D geometry within the broader paradigm of MLLM spatial reasoning. Rather than relying on specialized vision architectures or heavy dual-encoder designs, we highlight the camera pose as a lightweight signal that connects isolated frames into a continuous 3D space. By unifying continuous camera pose estimation and video understanding within a single MLLM, our proposed Cambrian-\textit{P} not only yields competitive streaming pose estimation but fundamentally endows the MLLM with a coherent, global understanding of the 3D physical world.

\section{\model}
\label{sec:model}

We introduce \model, a new video understanding paradigm for multimodal large language models by equipping it with native camera pose estimation capability. We start by introducing our framework in \cref{sec:architecture}, followed by the training objective and dynamics in \cref{sec:train_objective} and \cref{sec:train_dynamic}, respectively.

\begin{figure}[htbp]
  \centering
  \includegraphics[width=1\textwidth]{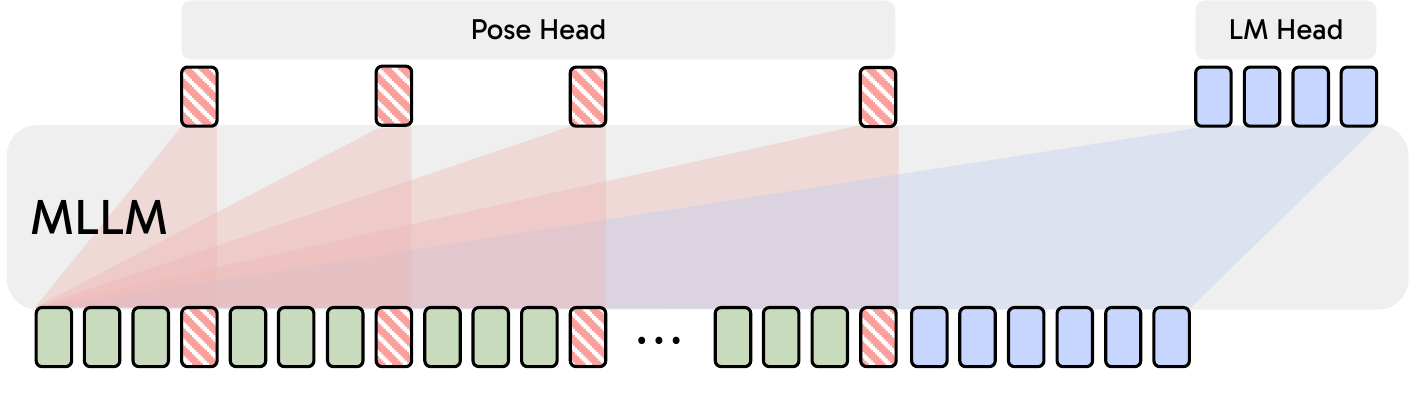}
    \caption{\textbf{\model Architecture Overview}. \model imposes minimal modifications to current MLLM architectures, introducing only {\setlength{\fboxsep}{1pt}\colorbox[HTML]{F3BFBC}{pose tokens instantiated from two learnable queries}} and a lightweight pose head. These tokens are marked with stripes. They are included during pose-supervised training and pose-estimation inference, but omitted by default for VQA inference. The pose tokens are appended to {\setlength{\fboxsep}{1pt}\colorbox[HTML]{CDDBC0}{visual tokens}}, while positioned before {\setlength{\fboxsep}{1pt}\colorbox[HTML]{C9D7FB}{text embeddings}}.}  \label{fig:architecture}
\end{figure}

\subsection{Architecture}
\label{sec:architecture}
As illustrated in \cref{fig:architecture}, our overall architecture introduces minimal additional components and overhead during both training and inference to enable camera pose estimation in the current MLLMs paradigm.

\noindent\textbf{MLLM.} We build \model upon the \cambrians~\cite{yang2026towards} architecture, a pretrained MLLM that pairs a SigLIP2-SO400m~\cite{tschannen2025siglip} vision encoder with a Qwen2.5~\cite{yang2024qwen2.5} LM connected via an MLP projector. 

\noindent\textbf{Camera Pose Tokens.} To enable camera pose estimation within the LLM's feature space, we introduce a small set of learnable camera pose tokens that are appended to each frame's visual tokens before they enter the LLM, inspired by the practice of VGGT~\cite{wang2025vggt}.
Specifically, we define two learnable queries $\mathbf{c}_{\text{first}}, \mathbf{c}_{\text{rest}} \in \mathbb{R}^{H}$, where $H$ is the LLM's hidden dimension. For a sequence of $N$ frames, we assign $\mathbf{c}_{\text{first}}$ to the first frame and $\mathbf{c}_{\text{rest}}$ to all remaining frames. This allows the model to distinguish the first frame from the rest, and to represent all poses in the coordinate system of the first camera. The per-frame token sequence fed to the LLM is:

\begin{equation}
[\,\mathbf{v}_i^{(1)},\ldots,\mathbf{v}_i^{(K)}\,;\,\mathbf{c}_i\,], \quad i = 1,\ldots,N,
\end{equation}

where $\mathbf{v}_i^{(j)}$ denotes the $K$ projected visual tokens, and $\mathbf{c}_i = \mathbf{c}_{\text{first}}$ for $i=1$, $ \mathbf{c}_i = \mathbf{c}_{\text{rest}}$ for $i>1$. Note that $\mathbf{c}_i$ is placed after the vision tokens of each frame due to the causal attention mechanism of the LLM.
After the LLM forwarding, we extract and slice out the pose token hidden state $\mathbf{h}_i \in \mathbb{R}^{H}$ for each frame from its final layer hidden states.

\noindent\textbf{Camera Pose Projector and Head.}
We bridge the LLM and the camera prediction head with a linear camera pose projector that maps LLM's hidden representation $\mathbf{h}_i$ to the required camera pose feature dimension as $\tilde{\mathbf{h}}_i = \mathbf{W}_p \mathbf{h}_i$.
To regress the camera parameters for each frame from $\{\tilde{\mathbf{h}}_i\}$, we adopt the camera head design of VGGT~\cite{wang2025vggt}, which includes four self-attention layers followed by a linear prediction layer.

\subsection{Training Objective}
\label{sec:train_objective}
Our training objective combines the next-token prediction loss for vision-language understanding with a camera pose estimation loss. The total loss is:

\begin{equation}
    \mathcal{L} = \mathcal{L}_{\text{NTP}} + \lambda_{\text{pose}} \cdot \mathcal{L}_{\text{pose}},
\end{equation}

where $\mathcal{L}_{\text{NTP}}$ is the standard cross-entropy loss over response text tokens, $\mathcal{L}_{\text{pose}}$ is the camera pose estimation loss, and $\lambda_{\text{pose}}$ is a weighting coefficient.

\noindent\textbf{Camera Pose Estimation Loss.}
Following VGGT~\cite{wang2025vggt}, we represent each camera as a pose encoding $\mathbf{g}_i = [\mathbf{t}_i, \mathbf{q}_i, f_i^h, f_i^w] \in \mathbb{R}^{9}$, where $\mathbf{t}_i \in \mathbb{R}^{3}$ is the absolute translation, $\mathbf{q}_i \in \mathbb{R}^{4}$ is the rotation quaternion, and $f_i^h, f_i^w \in \mathbb{R}$ encode the horizontal and vertical field-of-view. The camera pose loss supervises predicted pose encodings $\hat{\mathbf{g}}_i$ against ground truth $\mathbf{g}_i$ using a weighted L1 loss:
\begin{equation}
\mathcal{L}_{\text{pose}} = \frac{1}{N}\sum_{i=1}^{N} \left( \frac{w_T}{\bar{d}} \| s^* \hat{\mathbf{t}}_i - \mathbf{t}_i \|_1 + w_R \| \hat{\mathbf{q}}_i - \mathbf{q}_i \|_1 + w_f \| [\hat{f}_i^h,\hat{f}_i^w] - [f_i^h,f_i^w] \|_1 \right),
\end{equation}
where $w_T$, $w_R$, and $w_f$ are component weights and $\bar{d}$ is the trajectory-length normalization factor.

Following VGGT~\cite{wang2025vggt}, we canonicalize every ground-truth quaternion to the $w\!\geq\!0$ hemisphere before computing the loss, resolving the sign ambiguity that $\mathbf{q}$ and $-\mathbf{q}$ represent the same rotation. We do not explicitly normalize the predicted quaternion $\hat{\mathbf{q}}$ inside the L1 loss; supervision against the unit-norm ground truth implicitly encourages $\|\hat{\mathbf{q}}\|\!\to\!1$. For evaluation, the standard $\mathbf{q}\!\to\!R$ conversion is scale-invariant and includes a $1/\|\hat{\mathbf{q}}\|^2$ factor, so any non-zero predicted quaternion maps to a valid rotation matrix regardless of its magnitude. As training data can span a wide range of physical scales from indoor scenes to large outdoor driving sequences, the magnitude of translation errors can vary by orders of magnitude. Furthermore, non-metric datasets inherently possess arbitrary numerical scales, which would otherwise lead to unpredictable gradient magnitudes. To prevent large-scale scenes or arbitrarily scaled non-metric data from dominating the gradient, we normalize the translation loss term by the sequence-averaged consecutive frame distance of the ground-truth trajectory:
\begin{equation}
  \bar{d} = \frac{1}{N-1} \sum_{i=2}^{N} \| \mathbf{t}_i - \mathbf{t}_{i-1} \|_2,  
\end{equation}
which ensures that indoor and outdoor scenes contribute comparable gradients during training.

To include both metric-scale and non-metric-scale datasets~\cite{li2018megadepth,ling2024dl3dv} in training, we resolve the scale ambiguity inherent to non-metric data. Since the same camera trajectory can be encoded with any constant multiplier on all translations, its absolute scale is not physically meaningful. For non-metric samples, we compute a closed-form least-squares scale factor
$s^* = \operatorname{stop\_grad}\!\left(
\frac{\sum_i \hat{\mathbf{t}}_i \cdot \mathbf{t}_i}
{\sum_i \hat{\mathbf{t}}_i \cdot \hat{\mathbf{t}}_i}
\right),$
which rescales the predicted translations to the ground truth before the L1 loss, so the model is supervised on trajectory shape rather than arbitrary dataset scale. The stop-gradient on $s^*$ treats it as a constant during backpropagation; otherwise, the model could reduce the loss by collapsing $\hat{\mathbf{t}}\!\to\!0$ and letting $s^*$ absorb the trajectory scale. For metric-scale samples, we set $s^*=1$, so the absolute translation scale is directly supervised.

\subsection{Improving Training Dynamics}
\label{sec:train_dynamic}
While the architecture and training objective of \model are straightforward, the training dynamics present the most significant challenge when jointly optimizing VQA and camera pose estimation.

\noindent\textbf{Training Dynamics Gaps between VQA and Camera Pose Estimation.} The challenges primarily arise from three conflicts between their training paradigms. \textit{First}, a video frame sampling gap exists: MLLMs typically sample frames at fixed intervals regardless of the query. This yields repeated ground-truth poses across iterations, encouraging memorization of video-pose correspondences rather than genuine pose learning. In contrast, robust camera pose estimation requires random starting frames and dynamic temporal intervals in frame sampling~\cite{wang2025continuous,wang2025vggt,keetha2025mapanything}.
\textit{Second}, there is a gap in training duration. Advanced MLLMs typically train for only a single epoch, whereas pose estimation models require tens of epochs with diverse frame sampling to converge~\cite{wang2025continuous,wang2025vggt}.
\textit{Third}, the data augmentation gap complicates joint training. While VQA training generally omits augmentations to preserve the factual correctness of answers, camera pose estimation relies on augmentations like color jittering, Gaussian blur, and grayscale~\cite{wang2025vggt}. We empirically observe that applying these data augmentations to pose estimation samples is crucial for pose estimation and simultaneously benefits VQA performance.

\begin{figure}[t]
  \centering
  \includegraphics[width=\textwidth]{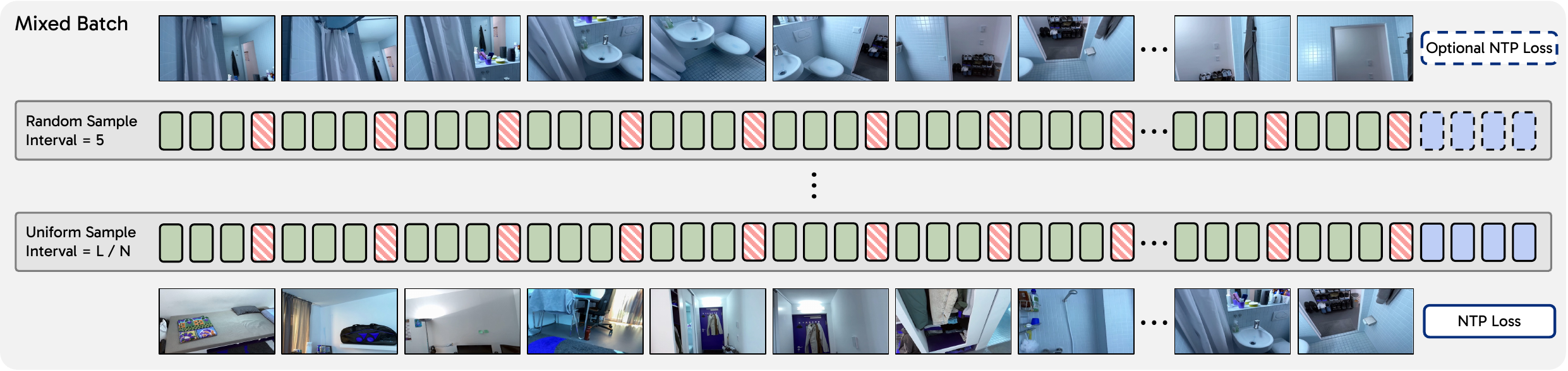}
  \caption{\textbf{Interleaved training of \model}. Top: augmented pose-only samples using dynamic frame sampling and only pose supervision. Bottom: samples using uniform frame sampling with both VQA and pose supervision. $L$ is the total number of frames of the video. }
  \label{fig:sampling}
\end{figure}

\noindent\textbf{Interleaved Training between VQA and Pose.} To overcome the aforementioned gaps, we introduce an interleaved training strategy with dedicated pose estimation samples that use their preferred sampling and augmentation strategy, and are supervised only by the pose loss. Specifically, given $\hat{M}$  training samples with pose supervision, we augment them by a ratio of $\beta$. The resulting $\lfloor \beta\hat{M} \rfloor$ augmented samples follow the standard sampling and augmentation strategies used in camera pose estimation and are trained with only the camera pose loss $\mathcal{L}_{\text{pose}}$. We omit the VQA loss here, as the limited temporal coverage of these samples lacks sufficient context for question answering (see \cref{fig:sampling}). Applying the VQA objective on incomplete visual information could encourage hallucination. 
Furthermore, the augmentation of pose-only samples enables us to arbitrarily scale training iterations for pose estimation, fully decoupled from the VQA objective. As shown in \cref{fig:sampling}, in our implementation, our batches are fully mixed, including samples with VQA-only, pose-only, or joint supervision.

\noindent\textbf{Random Jitter Frame Sampling.} In addition, for VQA samples loaded from scene-frame sequences with uniform sampling, we apply jitter augmentation to the sampled frame indices. Let $L$ denote the number of available frame entries in a sequence and $N$ the number of sampled frames, where $L\geq N$, with uniform indices $\{u_i\}_{i=1}^{N}$. We set $u_{N+1}=L$ and independently sample integer offsets $\delta_i \sim \operatorname{Unif}\{-\Delta,\ldots,\Delta\}$, where $\Delta=\lfloor L\alpha\rfloor$ and $\alpha$ controls the perturbation magnitude. We first form clipped proposals
\begin{equation}
r_i=\operatorname{clip}\!\left(u_i+\delta_i,\,0,\,u_{i+1}-1\right), \quad i=1,\ldots,N.
\end{equation}
The convention $u_{N+1}=L$ makes the last proposal clipped to $[0,L-1]$. We then enforce temporal order recursively:
\begin{equation}
\tilde{u}_1=r_1, \qquad \tilde{u}_i=\max\!\left(r_i,\tilde{u}_{i-1}\right), \quad i=2,\ldots,N.
\end{equation}
The final indices therefore satisfy $0\leq\tilde{u}_1\leq\cdots\leq\tilde{u}_N\leq L-1$. This strategy introduces temporal variability while preserving a valid frame order, alleviating memorization of fixed frame--pose correspondences under uniform sampling.

\noindent\textbf{Implementation Details.} We finetune \model from \cambrians-7B stage 3~\cite{yang2026towards} following its stage 4 training recipe. We perform end-to-end finetuning with AdamW optimizer with learning rates of $1 \times 10^{-5}$ for the LLM and vision projector, $2 \times 10^{-6}$ for the vision encoder, and $1 \times 10^{-4}$ for the pose projector and head. The pose projector and head are randomly initialized and trained from scratch. By default, we set the interleaved training augmentation ratio $\beta$ to $1$, the random jitter ratio $\alpha$ to $0.005$, and the loss trade-off factor $\lambda_\text{pose}$ to $0.2$. We train \model on 64 H200 GPUs with a 256 batch size.
For training data, we use \vsidata~\cite{yang2026towards} and data from MapAnything~\cite{keetha2025mapanything}. When only partial labels are available, \model activates only the corresponding loss, \ie, VQA loss or camera pose loss.

\section{Improved VQA with \model}
\subsection{Experiment Setups}

\noindent\textbf{Training Setups.} For fair comparison with \cambrians and existing MLLMs, we train \model with only data from \vsidata unless otherwise specified.

\noindent\textbf{Benchmarks.} We evaluate on a comprehensive suite of spatial reasoning and video understanding benchmarks, including VSI-Bench~\cite{yang2024think}, VSTI-Bench~\cite{fan2025vlm}, SPAR-Bench~\cite{zhang2025flatland}, MMSI-Bench~\cite{yang2025mmsi}, MMSI-Video-Bench~\cite{lin2025mmsi}, MindCube~\cite{yin2025spatial}, Tomato~\cite{shangguan2024tomato}, MVBench~\cite{li2024mvbench}, EgoSchema~\cite{mangalam2023egoschema}, and Perception Test~\cite{patraucean2023perception}. 

\noindent\textbf{Baselines.} We compare \model against three categories of models: (1)~general-purpose MLLMs including GPT-4o~\cite{hurst2024gpto}, Gemini 2.5 Pro~\cite{comanici2025gemini}, Qwen2.5-VL-7B~\cite{bai2025qwen2}, InternVL-3~\cite{zhu2025internvl3}, InternVL-3.5~\cite{wang2025internvl3}, and Qwen3-VL~\cite{Qwen3-VL}; (2)~spatial-specialist models including VST~\cite{yang2025visual}, VLM-3R~\cite{fan2025vlm}, VG-LLM~\cite{zheng2025learning}, SenseNova-SI~\cite{cai2025scaling}, \cambrians~\cite{yang2026towards}, and GeoThinker~\cite{li2026thinking}; and (3)~chance-level baselines (random and frequency).

\begin{table*}[!t]
\centering
\caption{\textbf{VSI-Bench Results Comparison}. $\dagger$ indicates this \cambrians is fine-tuned only on VSI-590K.}
\label{tab:vsi}
\resizebox{1\textwidth}{!}{%
\begin{tabular}{ll|c|cccc|cccc}
\toprule
\multirow{2}{*}{Model} & \multirow{2}{*}{LM} & & \multicolumn{4}{c|}{Numerical Answer} & \multicolumn{4}{c}{Multiple-Choice Answer} \\
\cmidrule(lr){4-7} \cmidrule(lr){8-11}
 & & Avg. & Obj. Count & Abs. Dist. & Obj. Size & Room Size & Rel. Dist. & Rel. Dir. & Route Plan & Appr. Order \\
\midrule
\rowcolor{gray!10!blue!5}\textit{Baselines} & & & & & & & & & &  \\
\textcolor{gray}{Chance Level (Random)} & \textcolor{gray}{--} & \textcolor{gray}{--} & \textcolor{gray}{--} & \textcolor{gray}{--} & \textcolor{gray}{--} & \textcolor{gray}{--} & \textcolor{gray}{25.0} & \textcolor{gray}{36.1} & \textcolor{gray}{28.3} & \textcolor{gray}{25.0} \\
\textcolor{gray}{Chance Level (Frequency)} & \textcolor{gray}{--} & \textcolor{gray}{34.0} & \textcolor{gray}{62.1} & \textcolor{gray}{32.0} & \textcolor{gray}{29.9} & \textcolor{gray}{33.1} & \textcolor{gray}{25.1} & \textcolor{gray}{47.9} & \textcolor{gray}{28.4} & \textcolor{gray}{25.2} \\
\midrule
\rowcolor{gray!10!blue!5}\textit{General-purpose Models} & & & & & & & & & &  \\
GPT-4o \cite{hurst2024gpto} & Unk. & 34.0 & 46.2 & 5.3 & 43.8 & 38.2 & 37.0 & 41.3 & 31.5 & 28.5 \\
Gemini 2.5 Pro \cite{comanici2025gemini} & Unk. & 51.5 & 43.8 & 34.9 & 64.3 & 42.8 & 61.1 & 47.8 & 45.9 & 71.3 \\
Qwen2.5-VL-7B \cite{bai2025qwen2} & Qwen2.5-7B & 29.3 & 25.2 & 10.5 & 36.4 & 29.6 & 38.4 & 38.0 & 29.8 & 26.8 \\
InternVL-3 8B \cite{zhu2025internvl3} & Qwen2.5-7B & 42.1 & 68.1 & 39.0 & 48.4 & 33.6 & 48.3 & 36.4 & 27.3 & 35.4 \\
InternVL-3.5 8B \cite{wang2025internvl3} & Qwen3-8B & 56.3 & -- & -- & -- & -- & -- & -- & -- & -- \\
Qwen3-VL 8B \cite{Qwen3-VL} & Qwen3-8B & 56.6 & -- & -- & -- & -- & -- & -- & -- & -- \\
\midrule
\rowcolor{gray!10!blue!5}\textit{Spatial-specialist Models} & & & & & & & & & &  \\
VST 7B \cite{yang2025visual} & Qwen2.5-7B & 61.2 & 71.6 & 43.8 & 75.5 & 69.2 & 60.0 & 55.6 & 44.3 & 69.2 \\
VLM-3R 7B \cite{fan2025vlm} & Qwen2-7B & 60.9 & 70.2 & 49.4 & 69.2 & 67.1 & 65.4 & 80.5 & 45.4 & 40.1 \\
VG-LLM 8B \cite{zheng2025learning} & Qwen2.5-7B & 50.7 & 67.9 & 37.7 & 58.6 & 62.0 & 46.6 & 40.7 & 32.4 & 59.2 \\
\cambrians 7B \cite{yang2026towards} & Qwen2.5-7B & 67.5 & 73.2 & 50.5 & 74.9 & 72.2 & 71.1 & 76.2 & 41.8 & 80.1 \\
SenseNova-SI 8B~\cite{cai2025scaling} & Qwen2.5-7B & 68.7 & -- & -- & -- & -- & -- & -- & -- & --\\
GeoThinker 7B \cite{li2026thinking} & Qwen2.5-7B & 68.5 & -- & -- & -- & -- & -- & -- & -- & -- \\
GeoThinker 8B \cite{li2026thinking} & Qwen3-8B & 72.6 & -- & -- & -- & -- & -- & -- & -- & -- \\
\cambrians-7B$^\dagger$~\cite{yang2026towards} & Qwen2.5-7B & 69.2 & 73.6 & 53.7 & 75.2 & 74.7 & 71.5 &  82.0 & 38.7 & 84.3 \\
\rowcolor{green!5} \textbf{\model} & Qwen2.5-7B & \textbf{73.7} & \textbf{74.9} & \textbf{60.1} & \textbf{76.0} & \textbf{76.9} & \textbf{74.8} & \textbf{89.5} & \textbf{52.6} & \textbf{85.0} \\
\bottomrule
\end{tabular}
}
\end{table*}

\subsection{Results}

\noindent\textbf{VSI-Bench.} As shown in \cref{tab:vsi}, \model yields state-of-the-art spatial reasoning capability in VSI-Bench. In particular, compared to existing spatial-specialist models such as \cambrians-7B, SenseNova-SI-8B, and GeoThinker-7B, \model achieves gains of 5.0--6.2 points.
Moreover, \model outperforms \cambrians$^\dagger
$, its counterpart without camera pose estimation, by 4.5 points, highlighting the effectiveness of incorporating camera pose prediction in spatial reasoning.
For per-subtask improvement, \model shows the most prominent improvement on absolute distance, relative direction, and route plan -- tasks that demand a more global understanding of the space. It is also noteworthy that \model shows superior out-of-distribution generalization capability in the route plan task, which is not included in the \vsidata training set, suggesting that \model learns beyond the exact task distribution.

\begin{table}[h]
\centering
\caption{\textbf{VSTI-Bench Result}. We finetune \model on \vsidata and VLM-3R~\cite{fan2025vlm} data for this experiment. \model shows a 20.0 percentage-point improvement on the camera movement direction subtask.}
\label{tab:vsti_results}
\vspace{-0.15cm}
\resizebox{\textwidth}{!}{%
\begin{tabular}{l|c|cc|ccc}
\toprule
Methods & Avg. & Cam-Obj Abs. Dist. & Cam. Displace. & Cam. Mov. Dir. & Obj-Obj Rel. Pos. & Cam-Obj Rel. Dist. \\
\midrule
GPT-4o \cite{hurst2024gpto}& 38.2 & 29.5 & 23.4 & 37.3 & 58.1 & 42.5 \\
Gemini-1.5 Flash \cite{team2024gemini} & 32.1 & 28.5 & 20.9 & 24.4 & 52.6 & 33.9 \\
LLaVA-NeXT-Video-72B \cite{zhang2024llavanextvideo} & 44.0 & 32.3 & 10.5 & 48.1 & 78.3 & 50.9 \\
VLM-3R-7B \cite{fan2025vlm} & 58.8 & 39.4 & 39.6 & 60.6 & 86.5 & 68.6 \\
GeoThinker 8B \cite{li2026thinking} & 67.4 & 38.4 & 45.8 & 84.2 & 93.6 & \textbf{75.2}\\
\midrule
\model\ (w/o Pose) & 62.4 & 39.4 & 40.6 & 67.7 & 92.2 & 72.0 \\
\model & \textbf{68.9} & \textbf{42.5} & \textbf{46.6} & \textbf{87.7} & \textbf{94.3} & 73.2 \\
\bottomrule
\end{tabular}%
}
\end{table}

\noindent\textbf{Significant Improvement in Understanding Camera Movement.} To further evaluate how well \model captures camera movement, we finetune it on \vsidata~\cite{yang2026towards} and VLM-3R~\cite{fan2025vlm} data and evaluate on VSTI-Bench~\cite{fan2025vlm}, which includes questions about camera motion. As shown in \cref{tab:vsti_results}, \model achieves state-of-the-art results on VSTI-Bench. More importantly, it obtains a 20.0 percentage-point improvement on the camera movement subtask over the no-pose baseline, demonstrating that the camera pose estimation objective directly enhances the model's understanding of camera dynamics.

\begin{table}[h]
\centering
\caption{\textbf{Out-of-Distribution Generalization for Spatial and General VQA Benchmarks}. \model is fine-tuned only on \vsidata, without any in-distribution training data for benchmarks here.}
\vspace{-0.15cm}
\label{tab:ood_vqa}
\resizebox{\textwidth}{!}{
\begin{tabular}{l|cccccccc}
\toprule
Model & SPAR-Bench & MMSI-Bench & MMSI-Video-Bench & MindCube & MVBench & EgoSchema & Perception Test & Tomato \\
\midrule
\model\ (w/o Pose) & 32.7 & 26.2 & 20.1 & 34.3 & 51.9 & 49.6 & 56.4 & 20.4 \\
\model             & \textbf{35.9} & \textbf{28.0} & \textbf{22.9} & \textbf{38.4} & \textbf{53.5} & \textbf{52.5} & \textbf{58.4} & \textbf{26.7} \\
\bottomrule
\end{tabular}
}
\end{table}

\noindent\textbf{OOD Improvement on Spatial and General VQA Benchmarks.} As shown in \cref{tab:ood_vqa}, although \model is finetuned solely on \vsidata, which is in-distribution with respect to VSI-Bench, it also demonstrates improvements on out-of-distribution spatial and general VQA benchmarks. This suggests that the local-to-global video understanding capability acquired through camera pose prediction is a general and fundamental skill transferable to broader video QA tasks.

\subsection{Improving General Video QA with Pseudo-Annotated Pose}
\model shows promising improvements on both spatial VQA and general VQA benchmarks when ground-truth pose supervision is available. However, GT camera poses are available only for limited data sources in \vsidata{} (\eg, ScanNet, ScanNet++, and ARKitScenes). To scale pose supervision to general-domain videos, we pseudo-annotate videos corresponding to the subsampled 590K samples from \cambrians{}-3M~\cite{yang2026towards}.
We curate pseudo poses using VIPE~\cite{huang2025vipe}. Video clips first pass a scene-cut detector and a quality filter based on Qwen3-VL~\cite{Qwen3-VL}. Remaining clips are processed by VIPE and post-filtered; see \cref{app:pseudo_pose_pipeline} for details. The resulting pseudo poses are used as GT poses in the interleaved training recipe.

As shown in \cref{tab:general_vqa}, adding general VQA data substantially improves general video QA performance on MVBench, Perception Test, and EgoSchema, but slightly degrades VSI-Bench. Introducing GT pose supervision reverses this spatial degradation, improving VSI-Bench while preserving the gains on general VQA. Adding pseudo-pose supervision from general-domain videos further boosts all four benchmarks, yielding additional gains on VSI-Bench as well as substantial improvements on MVBench and EgoSchema. These results suggest that pseudo poses, even when derived from noisy in-the-wild videos, provide a scalable supervision signal for video understanding.

\begin{table}[th]
\centering
\caption{\textbf{\model with general VQA training data and pseudo pose supervision.}
We report \model 128-frame results on VSI-Bench, MVBench, Perception Test, and EgoSchema.}
\label{tab:general_vqa}
\resizebox{0.98\textwidth}{!}{%
\begin{tabular}{l l c cccc}
\toprule
\textbf{Training Data} & \textbf{Pose Sup.} & \textbf{\% Pose Sup.} & \textbf{VSI-Bench} & \textbf{MVBench} & \textbf{Perception Test} & \textbf{EgoSchema} \\
\midrule
\rowcolor{gray!10!blue!5}\textit{Spatial VQA Data Only} & & & & & & \\
VSI-590K            & --          & 0\%   & 71.2 & 51.7 & 56.7 & 48.5 \\
VSI-590K            & GT          & 49\%       & 73.7 & 53.8 & 58.1 & 51.3 \\
\midrule
\rowcolor{gray!10!blue!5}\textit{Spatial VQA + General VQA Data} & & & & & & \\
VSI-590K + CamS-590K & --            & 0\%  & 70.9 & 68.0 & 66.9 & 71.2 \\
VSI-590K + CamS-590K & GT            & 25\%   & 73.7 & 67.9 & 67.8 & 71.7 \\
VSI-590K + CamS-590K & GT + Pseudo   &  48\% & \textbf{73.9} & \textbf{69.3} & \textbf{67.9} & \textbf{73.6} \\
\bottomrule
\end{tabular}}
\end{table}

\section{Camera Pose Estimation with \model}
\label{sec:pose_exp}
\subsection{Experiment Setups}
\noindent\textbf{Training Setups.} To further push the camera pose estimation capability, we train \model on data with pose annotation from \vsidata (\ie, ScanNet~\cite{dai2017scannet}, ScanNet++~\cite{yeshwanth2023scannet++}, and ARKitScenes~\cite{dehghan2021arkitscenes}) and datasets from MapAnything~\cite{keetha2025mapanything}, which include metric-scale datasets (ParallelDomain4D \cite{van2024generative}, TartanAir-v2~\cite{wang2020tartanair,zhang2025ufm}, MVS-Synth~\cite{huang2018deepmvs}, Spring~\cite{mehl2023spring}, SailVOS3D~\cite{hu2021sail}, ETH3D~\cite{schops2017multi}, Dynamic Replica~\cite{karaev2023dynamicstereo}, MPSD \cite{antequera2020mapillary}, and UnrealStereo4K~\cite{tosi2021smd}) and non-metric-scale datasets (MegaDepth~\cite{li2018megadepth}, DL3DV~\cite{ling2024dl3dv}, and BlendedMVS~\cite{yao2020blendedmvs}). To further boost the performance on camera pose estimation, we set the interleaved training augmentation ratio $\beta$ to $20$ and the loss trade-off factor $\lambda_\text{pose}$ to $0.5$.

\begin{table*}[tb]
\centering
\caption{\textbf{Camera pose estimation results on ScanNet, TUM, and Sintel}. \model is trained on \vsidata and MapAnything data to improve camera pose estimation capability.}
\label{tab:main_result_pose}
\resizebox{1\textwidth}{!}{
\begin{tabular}{l|ccc|ccc|ccc}
\toprule
\multirow{2}{*}{Model} & \multicolumn{3}{c|}{\textbf{ScanNet}} & \multicolumn{3}{c|}{\textbf{TUM-dynamic}} & \multicolumn{3}{c}{\textbf{Sintel}} \\
& ATE $\downarrow$ & RPE trans $\downarrow$ & RPE rot $\downarrow$ & ATE $\downarrow$ & RPE trans $\downarrow$ & RPE rot $\downarrow$ & ATE $\downarrow$ & RPE trans $\downarrow$ & RPE rot $\downarrow$ \\
\midrule
\rowcolor{gray!10!blue!5}\textit{Offline Models} & & & & & & & & & \\
VGGT \cite{wang2025vggt} & 0.035 & 0.015 & 0.380 & \textbf{0.009} & \textbf{0.008} & 0.350 & 0.172 & 0.061 & 0.470 \\
DUSt3R-GA \cite{wang2024dust3r} & 0.081 & 0.028 & 0.784 & 0.083 & 0.017 & 3.567 & 0.417 & 0.250 & 5.796 \\
MASt3R-GA \cite{leroy2024grounding} & 0.078 & 0.020 & 0.475 & 0.038 & 0.012 & 0.448 & 0.185 & 0.060 & 1.496 \\
MonST3R-GA \cite{zhang2025monstr} & 0.077 & 0.018 & 0.529 & 0.098 & 0.019 & 0.935 & 0.111 & 0.044 & 0.869 \\
Fast3R \cite{yang2025fast3r} & 0.155 & 0.123 & 3.491 & 0.090 & 0.101 & 1.425 & 0.371 & 0.298 & 13.750 \\
FLARE \cite{zhang2025flare} & 0.064 & 0.023 & 0.971 & 0.026 & 0.013 & 0.475 & 0.207 & 0.090 & 3.015 \\
$\pi^3$ \cite{wang2026pi} & \textbf{0.031} & \textbf{0.013} & \textbf{0.347} & 0.014 & 0.009 & \textbf{0.312} & \textbf{0.074} & \textbf{0.040} & \textbf{0.282} \\
MapAnything \cite{keetha2025mapanything} & 0.052 & 0.025 & 0.720 & 0.029 & 0.023 & 0.370 & 0.226 & 0.077 & 0.640 \\
\midrule
\rowcolor{gray!10!blue!5}\textit{Streaming Models} & & & & & & & & &  \\
StreamVGGT \cite{zhuo2026streaming} & 0.127 & 0.041 & 1.880 & 0.062 & 0.030 & 0.690 & 0.273 & 0.109 & 0.850 \\
CUT3R \cite{wang2025continuous} & 0.096 & \textbf{0.022} & \textbf{0.590} & \textbf{0.045} & \textbf{0.015} & \textbf{0.440} & \textbf{0.215} & \textbf{0.070} & \textbf{0.630} \\
Point3R \cite{wu2025point3r} & 0.097 & 0.035 & 2.791 & 0.058 & 0.031 & 0.758 & 0.442 & 0.154 & 1.897 \\
Spann3R \cite{wang20253d} & 0.096 & 0.023 & 0.661 & 0.056 & 0.021 & 0.591 & 0.329 & 0.110 & 4.471 \\
G$^2$VLM \cite{hu2025g} & 0.148 & 0.048 & 1.220 & 0.129 & 0.044 & 0.700 & 0.301 & 0.135 & 1.450 \\
\rowcolor{green!5} \textbf{\model} & \textbf{0.078} & 0.023 & 0.880 & 0.046 & 0.020 & 0.580 & 0.239 & 0.081 & 2.440 \\
\bottomrule
\end{tabular}
}
\end{table*}

\noindent\textbf{Benchmarks.} We evaluate camera pose estimation on three benchmarks: ScanNet~\cite{dai2017scannet}, TUM-dynamic~\cite{sturm2012benchmark}, and Sintel~\cite{butler2012naturalistic}, covering indoor scenes, handheld sequences, and synthetic movies with camera motions. Following MonST3R~\cite{zhang2025monstr}, for TUM-dynamic and ScanNet, we sample the first 90 frames with a temporal stride of 3, and for Sintel, we exclude static scenes or sequences with near-straight camera motion. Although the default training setup uses 32-frame sequences, \model's streaming architecture accepts arbitrary-length sequences; Table~\ref{tab:main_result_pose} therefore evaluates 90-frame sequences at inference. We report three metrics: Absolute Trajectory Error (ATE), Relative Pose Error in translation (RPE trans), and Relative Pose Error in rotation (RPE rot). All metrics are computed with Sim(3) alignment.

\noindent\textbf{Baselines.}
We compare \model against two categories of methods. Offline methods that require access to all frames simultaneously include VGGT~\cite{wang2025vggt}, DUSt3R~\cite{wang2024dust3r}, MASt3R~\cite{leroy2024grounding}, MonST3R~\cite{zhang2025monstr}, Fast3R~\cite{yang2025fast3r}, FLARE~\cite{zhang2025flare}, $\pi^3$~\cite{wang2026pi}, and MapAnything~\cite{keetha2025mapanything}, all evaluated with global alignment (GA) where applicable. Streaming methods that process frames incrementally include StreamVGGT~\cite{zhuo2026streaming}, CUT3R~\cite{wang2025continuous}, Point3R~\cite{wu2025point3r}, Spann3R~\cite{wang20253d}, and G$^2$VLM~\cite{hu2025g}.

\subsection{Results}
As shown in \cref{tab:main_result_pose}, \model achieves the lowest ATE on ScanNet~\cite{dai2017scannet} among streaming camera pose estimation models and delivers competitive performance on TUM-dynamic~\cite{sturm2012benchmark} and Sintel~\cite{butler2012naturalistic}, without relying on specialized designs like DINOv2 encoder~\cite{oquab2023dinov2} or bidirectional transformer~\cite{wang2025continuous,wang2025vggt}. This highlights that standard MLLMs can predict accurate camera pose with only an additional pose head and two learnable pose queries.
In addition, benefiting from the compact representation of the SigLIP encoder~\cite{tschannen2025siglip}, the lower FLOPs of the causal transformer, and the optimized inference infrastructure of the LLM ecosystem, \model shows competitive latency despite its large model size; see \cref{app:latency_details} for additional analysis.

\section{Scaling \model with Model, Data, and Training Steps}
The remarkable success of LLMs and the next-token prediction paradigm can be largely attributed to their scalability. We investigate whether the camera pose estimation objective exhibits similar scaling behavior within the MLLM paradigm, across model size, data size, and training iterations.

\begin{figure}[th]
    \centering
    \begin{subfigure}[t]{0.49\textwidth}
        \centering
        \includegraphics[width=\textwidth]{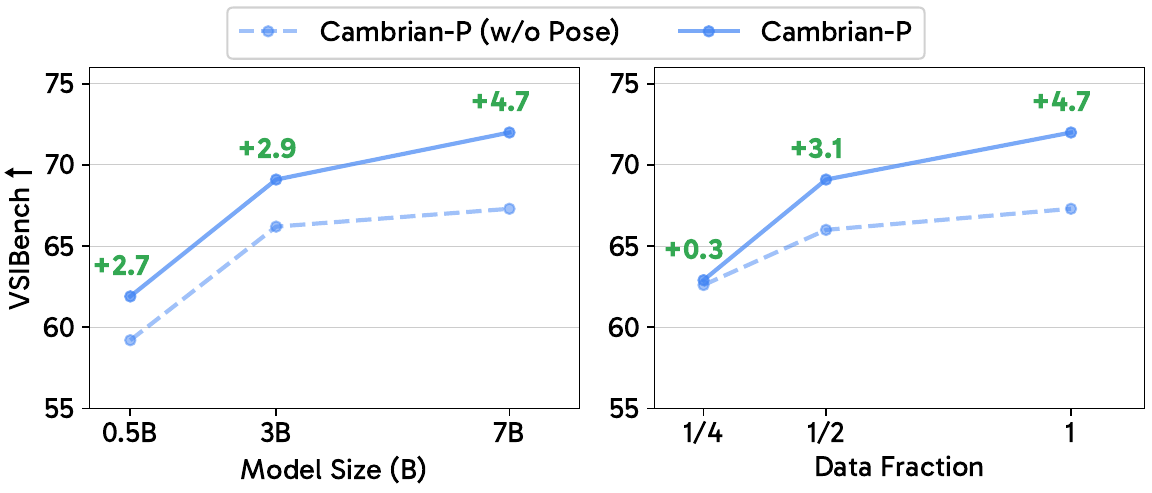}
        \caption{VQA acc. with various models and data sizes.}
        \label{fig:scaling_vqa}
    \end{subfigure}
    \hfill
    \begin{subfigure}[t]{0.49\textwidth}
        \centering
        \includegraphics[width=\textwidth]{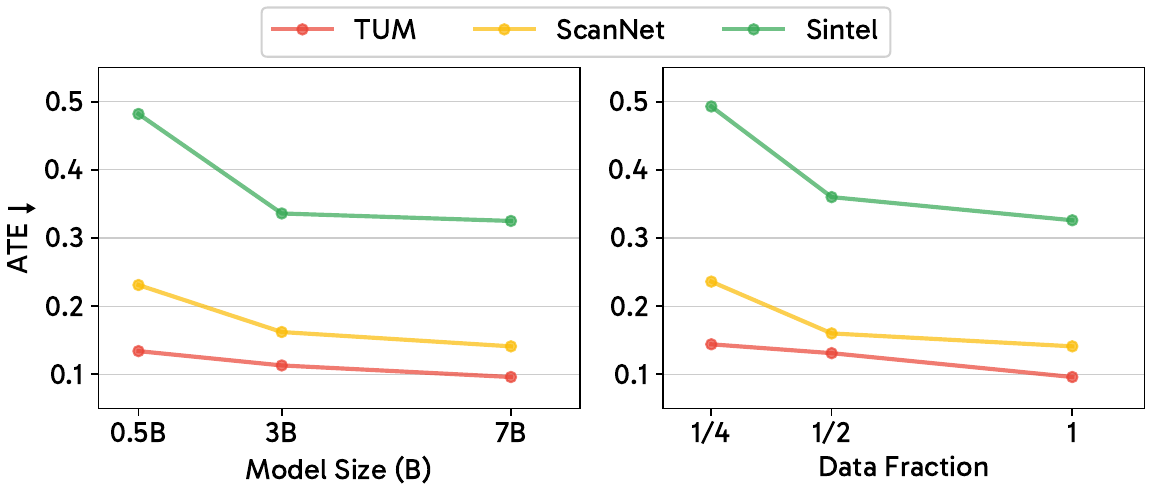}
        \caption{Pose ATE with various models and data sizes.}
        \label{fig:scaling_pose}
    \end{subfigure}
    \caption{\textbf{Comparison of \model regarding different model size and data size}. Larger models or more data both yield higher VSI-Bench scores and lower pose estimation error across all benchmarks.}
    \label{fig:scaling_combined}
\end{figure}

\noindent\textbf{Model Size Scaling.} To investigate the effect of model size, we finetune \model from \cambrians variants with different LLM sizes. As shown in \cref{fig:scaling_vqa}, scaling up the model not only improves VSI-Bench performance but also widens the gap over the no-pose baseline. We attribute this trend to the inherent demands of multi-task learning, where larger model capacity better accommodates the additional complexity. As shown in \cref{fig:scaling_pose}, the ATE of camera pose decreases as the model size increases.

\noindent\textbf{Data Size Scaling.} As shown in \cref{fig:scaling_vqa}, scaling up data size improves VSI-Bench performance while widening the gap over the no-pose baseline in the 7B model. Note that \model yields only marginal improvement with $\frac{1}{4}$ data, likely because the pose head is trained from scratch and struggles to converge with limited supervision. We empirically find that pretraining the pose head can alleviate this issue. Also, as shown in \cref{fig:scaling_pose}, the translation error of the camera pose consistently decreases with larger data size.

\noindent\textbf{Training Iteration Scaling.} As shown in \cref{tab:scale_iter_vqa}, scaling the augmented pose iterations in interleaved training is more efficient and scalable than increasing VQA iterations for improving VQA performance. Even without extra pose training iterations from interleaved training, adding pose supervision still yields a 2.1-point improvement. For scaling training iterations to improve camera pose estimation, we adopt the experimental setup described in \cref{sec:model}. As shown in \cref{tab:scale_iter_pose}, increasing the pose iterations leads to consistent decreases in ATE across all three pose benchmarks, demonstrating good scalability of MLLMs for camera pose estimation. Moreover, even when trained on a large amount of out-of-distribution data and with the pose estimation objective dominating, \model still achieves improved VQA performance on VSI-Bench, highlighting the synergy between spatial VQA and camera pose estimation.

\begin{table}[ht]
\centering
\caption{\textbf{Scaling training iterations to improve VQA}. We compare the performance of \model with and without pose supervision among different training iterations.}
\label{tab:scale_iter_vqa}
\resizebox{1\textwidth}{!}{%
\begin{tabular}{lcc|cccc}
\toprule
Model & VQA Iteration & Pose Iteration & VSI-Bench $\uparrow$ & ScanNet ATE$\downarrow$ & TUM ATE $\downarrow$ & Sintel ATE $\downarrow$ \\
\midrule
\multirow{3}{*}{\model (w/o Pose)} & 2K & 0 & 67.3 & - & - & - \\
         & 4K & 0 & 69.3 & - & - & - \\
         & 6K & 0 & 69.3 & - & - & - \\
\midrule
\multirow{5}{*}{\model} & 2K & 0 & 69.4 & 0.259 & 0.132 & 0.521 \\
     & 0 & 2K & 25.2 & 0.163 &	0.115	& 0.374 \\
     & 2K & 1K & 72.0 & \textbf{0.141} & \textbf{0.096} & \textbf{0.325} \\
     & 2K & 2K & 72.2 & 0.149 & 0.112 & 0.329 \\
     & 4K & 2K & \textbf{72.7} & 0.143 & 0.106 & 0.406 \\
\bottomrule
\end{tabular}}
\end{table}

\begin{table}[ht]
\centering
\caption{\textbf{Scaling pose iterations with MapAnything data}. \model is trained with both \vsidata and MapAnything data.}
\label{tab:scale_iter_pose}
\resizebox{\textwidth}{!}{%
\begin{tabular}{lcc|cccc}
\toprule
Model & VQA Iteration & Pose Iteration & VSI-Bench $\uparrow$ & ScanNet ATE$\downarrow$ & TUM ATE $\downarrow$ & Sintel ATE $\downarrow$ \\
\midrule
\model (w/o Pose) & \color{gray} 2K & \color{gray} 0 & \color{gray} 67.3 & \color{gray} -& \color{gray} -& \color{gray} - \\
\midrule
\multirow{4}{*}{\model} & 2K & 1K & \textbf{71.6} & 0.145 & 0.105 & 0.361 \\
     & 2K & 3K & 70.9 & 0.106 & 0.073 & 0.297 \\
     & 2K & 5K & 69.8 & 0.094 & 0.071 & 0.289 \\
     & 2K & 20K & 69.3 & \textbf{0.077} & \textbf{0.048} & \textbf{0.278} \\
\bottomrule
\end{tabular}}
\end{table}

\section{Analysis}

To better understand the property of \model, we analyze its behavior through extensive experiments covering component ablations, frame scaling, loss design, and qualitative trends. Without other specifications, all \model variants are trained with \vsidata in 32 frames and 196 tokens per frame.

\subsection{Ablation Studies}

\begin{table}[h]
\centering
\caption{\textbf{Components ablation study of \model}.}
\label{tab:ablation_component}
\resizebox{\textwidth}{!}{
\begin{tabular}{ccc|cccc}
\toprule
Camera Loss & Interleaved Training & Random Jitter & VSI-Bench $\uparrow$ & ScanNet ATE$\downarrow$ & TUM ATE $\downarrow$ & Sintel ATE $\downarrow$ \\
\midrule
\color{gray} &\color{gray}  &\color{gray}  & \color{gray}67.3 & \color{gray}- & \color{gray}- & \color{gray}- \\
\cmark & \cmark &  & 71.2 & 0.144 & 0.106 & 0.366 \\
\cmark &  & \cmark & 69.4 & 0.259 & 0.132 & 0.521 \\
\cmark & \cmark & \cmark & \textbf{72.0} & \textbf{0.141} & \textbf{0.096} & \textbf{0.325} \\
\bottomrule
\end{tabular}
}
\end{table}

\noindent\textbf{Component Ablation} As shown in \cref{tab:ablation_component}, pose loss and interleaved training yield a 3.9-point improvement on VSI-Bench while substantially reducing the ATE of camera pose estimation. Incorporating random jitter further brings a 0.8-point gain on VQA, along with better pose accuracy. These results suggest that both interleaved training and random jitter significantly mitigate the training dynamics gaps between the VQA and camera pose objectives.

\begin{table}[tb]
\centering
\caption{\textbf{Ablation on the number of input frames during training}.}
\label{tab:vsi_frame}
\resizebox{0.9\textwidth}{!}{%
\begin{tabular}{lccccc}
\toprule
Model & \# frames / \# tok & VSI-Bench$\uparrow$ & ScanNet ATE $\downarrow$ & TUM ATE $\downarrow$ & Sintel ATE $\downarrow$ \\
\midrule
{\color{gray}\model (w/o Pose)} & \multirow{2}{*}{32 / 196} & {\color{gray}67.3} & {\color{gray}--} & {\color{gray}--} & {\color{gray}--} \\
\model & & 72.0\rlap{ {\color{green!60!black}(+4.7)}} & 0.141 & 0.096 & 0.325 \\
\midrule
{\color{gray}\model (w/o Pose)} & \multirow{2}{*}{64 / 64} & {\color{gray}70.3} & {\color{gray}--} & {\color{gray}--} & {\color{gray}--} \\
\model & & 73.1\rlap{ {\color{green!60!black}(+2.8)}} & 0.140 & 0.104 & 0.272 \\
\midrule
{\color{gray}\model (w/o Pose)} & \multirow{2}{*}{128 / 64} & {\color{gray}71.2} & {\color{gray}--} & {\color{gray}--} & {\color{gray}--} \\
\model & & 73.7\rlap{ {\color{green!60!black}(+2.5)}} & 0.141 & 0.111 & 0.322 \\
\bottomrule
\end{tabular}%
}
\end{table}

\noindent\textbf{Number of Frames Ablation.} As shown in \cref{tab:vsi_frame}, \model achieves higher VQA performance on VSI-Bench as the number of input frames increases, while the gap over the baseline shrinks accordingly. We attribute this trend to VQA benefiting more from additional frames than  pose estimation does: more frames provide richer visual context for answering questions, while pose estimation learns better with lower inter-frame overlap and is typically trained on sequences of only 12$\sim$24 frames~\cite{wang2025vggt}. This is further supported by the ATE results on pose benchmarks, which slightly degrade as the frame count increases.

\subsection{How Does Camera Pose Help Video QA?}
Here, to understand how camera pose supervision helps, we investigate individual loss components and analyze VQA accuracy across varying spatial distances.

\begin{table}[h]
\centering
\caption{\textbf{Effect of pose tokens during training and inference in \model}. \model's improvements are driven by pose supervision during training, not by pose-token conditioning at inference.}
\label{tab:with_without_pose_vqa}
\resizebox{\textwidth}{!}{
\begin{tabular}{cc|cccccc}
\toprule
\multicolumn{2}{c|}{Pose Token} & \multirow{2}{*}{VSI-Bench} & \multirow{2}{*}{VSTI-Bench} & \multirow{2}{*}{MVBench} & \multirow{2}{*}{EgoSchema} & \multirow{2}{*}{Perception Test} & \multirow{2}{*}{Tomato} \\
Training & Inference & & & & & & \\
\midrule
\color{gray} \xmark & \color{gray} \xmark & \color{gray} 67.3 & \color{gray} 55.1 & \color{gray} 51.9 & \color{gray} 49.6 & \color{gray} 56.4 & \color{gray} 20.4 \\
\cmark & \cmark & \textbf{72.0} & 56.5 & \textbf{53.5} & \textbf{52.5} & 58.4 & \textbf{26.7} \\
\cmark & \xmark & \textbf{72.0} & \textbf{56.6} & 53.2 & \textbf{52.5} & \textbf{58.8} &\textbf{26.7} \\
\bottomrule
\end{tabular}
}
\end{table}

\noindent\textbf{Camera Pose Helps MLLMs Learn.} Starting from a standard MLLM, \model introduces pose tokens to leverage pose supervision during training and conditions on these tokens at inference. A natural question is whether the improvement on Video QA stems from the estimated pose trajectory provided at inference time, or from learning better representations through pose supervision during training. 
As shown in \cref{tab:with_without_pose_vqa}, pose tokens and supervision during training yield significant gains across various video benchmarks, whereas conditioning on pose tokens at inference time provides no additional benefit. This indicates that \model's improvements stem from the stronger representations learned under pose supervision during training, rather than from pose conditioning at inference.

\begin{table}[th]
\centering
\caption{\textbf{Loss ablation study results}. T, R, and FV indicate translation, rotation, and field-of-view loss.}
\label{tab:loss_ablation}
\resizebox{0.85\textwidth}{!}{%
\begin{tabular}{lccccc}
\toprule
Pose Loss & Depth Loss & VSI-Bench $\uparrow$ & ScanNet ATE$\downarrow$ & TUM ATE $\downarrow$ & Sintel ATE $\downarrow$ \\
\midrule
\color{gray}\xmark & \color{gray}\xmark & \color{gray}67.3 & \color{gray}- & \color{gray}- & \color{gray}- \\
\cmark & \xmark & \textbf{72.0} & \textbf{0.141} & 0.096 & 0.325 \\
\cmark & \cmark & 71.7 & 0.156 & 0.118 & \textbf{0.318} \\
\xmark & \cmark & 69.4 & 0.371 & 0.194 & 0.537 \\
T only & \xmark & 70.7 & 0.205 & 0.128 & 0.357 \\
R only & \xmark & 69.7 & 0.287 & \textbf{0.092} & 0.353 \\
FV only & \xmark & 69.4 & 0.408 & 0.196 & 0.670 \\
T + R & \xmark & 71.5 & 0.165 & 0.130 & 0.386 \\
\bottomrule
\end{tabular}}
\end{table}

\noindent\textbf{Camera Pose Helps VQA More than Depth.} To study the effect of depth supervision,
we attach a modified VGGT depth head that incorporates RMSNorm layers. We adopt the weighting factor from VGGT to balance the depth and pose losses. As shown in \cref{tab:loss_ablation}, adding pose loss alone improves VSI-Bench accuracy by 2.6 points over depth loss. Combining both losses leads to a slight degradation in both VQA and camera pose estimation over the pose loss alone baseline. This indicates that camera pose estimation has greater synergy with video understanding as probed by VQA. 
We attribute the underperformance of depth supervision to two factors: (i) predicting dense per-pixel depth from only 196 or 64 visual tokens makes multi-task optimization difficult; (ii) VGGT's depth supervision is local and, unlike pose, provides no global scene understanding.
Breaking down the pose loss into its components, we find that both translation and rotation losses effectively improve VQA performance, while field-of-view loss yields gains comparable to those from depth loss. We provide detailed setup ablation studies for incorporating depth supervision in \cref{app:depth_fairness}.

\begin{figure}[htbp]
  \centering
  \includegraphics[width=0.85\textwidth]{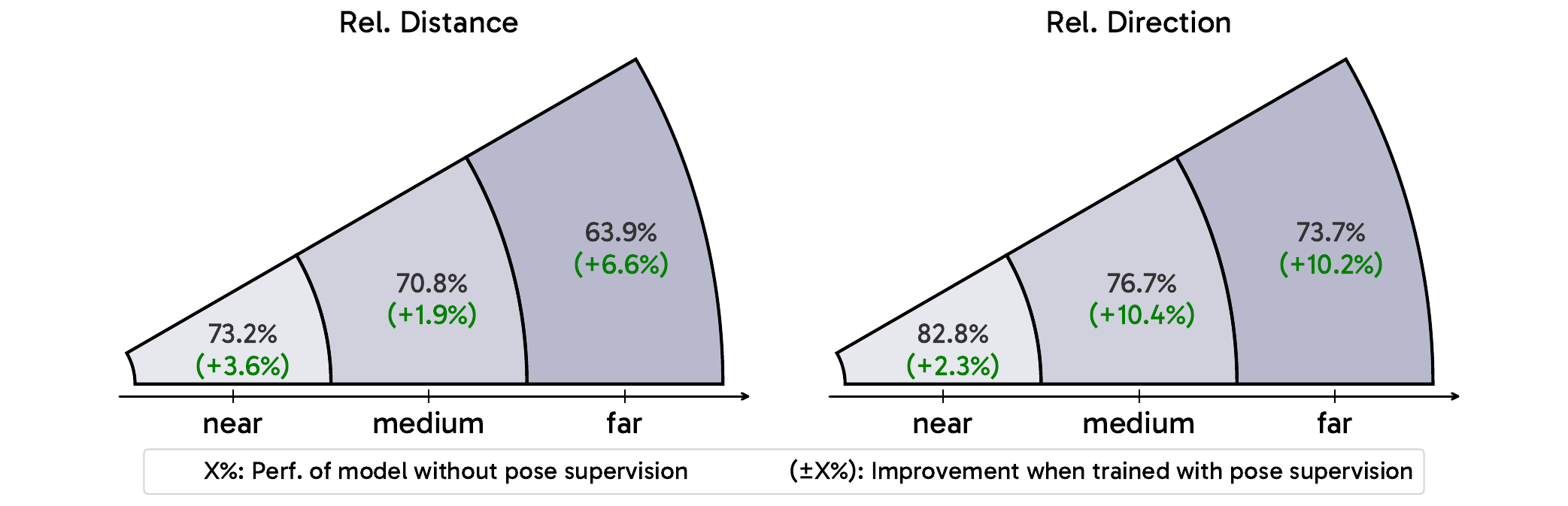}
  \caption{
    \textbf{
        Camera pose improves global spatial reasoning.
    }
    We first normalize the ground-truth distance by room size, and then use \texttt{np.geomspace} to group the samples into 3 groups (near, medium, and far), which are equally spaced on the log scale.
    The near/medium/far sample proportions are 15.8\%/66.9\%/17.3\% for \texttt{Rel.~Dist.} and 9.1\%/64.3\%/26.6\% for \texttt{Rel.~Dir.}, respectively.
  }
  \label{fig:close_to_far_improvement}
\end{figure}

\noindent\textbf{Camera Pose Enables More Global Spatial Reasoning.} VSI-Bench~\cite{yang2024think} observes that MLLMs fall short in spatial intelligence as they tend to see locally rather than globally. Here, we investigate whether enabling the MLLM to be aware of camera movement facilitates more global spatial reasoning. As shown in \cref{fig:close_to_far_improvement}, we group samples based on normalized ground-truth distance relative to room size into near, medium, and far categories for the relative distance and relative direction question types in VSI-Bench. We find that without pose supervision, model performance degrades as objects become farther apart, while \model exhibits larger gains for distant objects compared to nearby ones. This indicates that camera pose supervision enables MLLMs to develop more global spatial reasoning capabilities.

\begin{table*}[h]
\centering
\caption{\textbf{\model results finetuned from different \cambrians variants}. CamS-S1, S2, S3 represent checkpoints from increasing training stages of \cambrians.}
\label{tab:vqa_help_pose}
\resizebox{\textwidth}{!}{%
\begin{tabular}{lcccc|ccc}
\toprule
Model & VSI-Bench $\uparrow$ & EgoSchema $\uparrow$ & Percept. Test $\uparrow$ & MVBench $\uparrow$ & ScanNet ATE $\downarrow$ & TUM ATE $\downarrow$ & Sintel ATE $\downarrow$ \\
\midrule
\textcolor{gray}{CamS-S1} & \textcolor{gray}{21.4} & \textcolor{gray}{42.9} & \textcolor{gray}{44.4} & \textcolor{gray}{43.9} & \textcolor{gray}{-} & \textcolor{gray}{-} & \textcolor{gray}{-} \\
\model~(FT CamS-S1) & 68.1 & - & - & - & 0.130 & 0.085 & 0.366 \\
\midrule
\textcolor{gray}{CamS-S2} & \textcolor{gray}{24.6} & \textcolor{gray}{47.5} & \textcolor{gray}{53.5} & \textcolor{gray}{49.2} & \textcolor{gray}{-} & \textcolor{gray}{-} & \textcolor{gray}{-} \\
\model~(FT CamS-S2) & 69.6 & - & - & - & 0.105 & 0.073 & 0.285 \\
\midrule
\textcolor{gray}{CamS-S3} & \textcolor{gray}{35.7} & \textcolor{gray}{76.9} & \textcolor{gray}{70.8} & \textcolor{gray}{66.3} & \textcolor{gray}{-} & \textcolor{gray}{-} & \textcolor{gray}{-} \\
\model~(FT CamS-S3) & \textbf{69.8} & - & - & - & \textbf{0.094} & \textbf{0.071} & \textbf{0.289} \\
\bottomrule
\end{tabular}%
}
\end{table*}

\subsection{Can Video QA Help Camera Pose Estimation?}
We have extensively discussed how the 3D prior from camera pose estimation benefits video QA. But does the reverse also hold---can VQA improve camera pose estimation? As shown in \cref{tab:vqa_help_pose}, when the pretrained MLLM is more grounded in video QA in terms of better VSI-Bench~\cite{yang2024think}, EgoSchema~\cite{mangalam2023egoschema}, Perception Test~\cite{patraucean2023perception}, and MVBench~\cite{li2024mvbench} performance, the model finetuned on MapAnything~\cite{keetha2025mapanything} data predicts more accurate camera poses. We attribute this to the better video-language alignment via VQA pretraining, which provides a more effective foundation for the post-LLM camera pose head.

\subsection{Qualitative Results}
\begin{figure}[h]
  \centering
  \includegraphics[width=\textwidth]{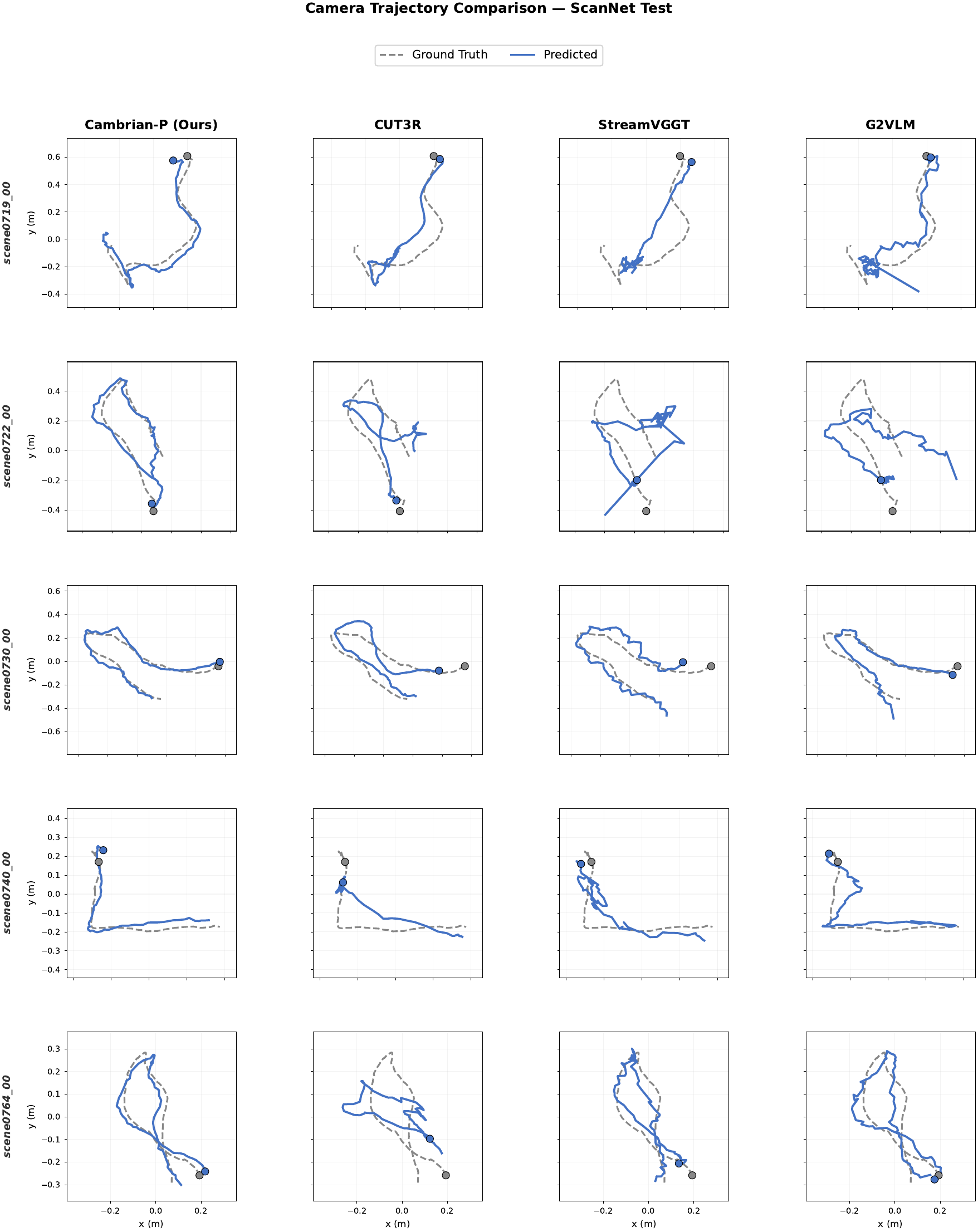}
  \vspace{-0.5cm}
  \caption{\textbf{Camera pose trajectory visualization on ScanNet test scenes.} These scenes are disjoint from the VSI-Bench evaluation sequences. \model generalizes well to unseen indoor environments.}
  \label{fig:scannet_traj_test}
\end{figure}

We show qualitative camera pose trajectory comparisons on ScanNet. For each scene, we plot the ground-truth trajectory (gray dashed) alongside predictions (blue solid) from \model, CUT3R~\cite{wang2025continuous}, StreamVGGT~\cite{zhuo2026streaming}, and G$^2$VLM~\cite{hu2025g}. All predicted trajectories are aligned to the ground truth via Sim(3) alignment and projected onto the two axes of greatest spatial extent for visualization.
\cref{fig:scannet_traj_test} shows five scenes from the ScanNet test split. \model generalizes well to these unseen indoor environments, maintaining accurate trajectory shapes across diverse room layouts and camera motions. 

Additional qualitative results on ScanNet validation scenes are provided in \cref{app:scannet_val_visualization}. We further visualize OOD predicted trajectories on EgoSchema clips in \cref{app:ood_pose_egoschema}, where \model is compared against specialist pose models using VIPE pseudo-GT trajectories. Qualitative examples in \cref{app:vqa_viz} illustrate how pose supervision helps \model answer spatial questions.

\subsection{Latency Analysis}
Although \model contains more parameters than specialist 3D reconstruction models, it remains efficient for camera pose estimation due to its compact visual representation, causal transformer backbone, and optimized LLM inference stack. We benchmark latency against recent specialist models~\cite{wang2025vggt,wang2025continuous,zhuo2026streaming} on the ScanNet~\cite{dai2017scannet} test set with a single NVIDIA L40S GPU, excluding data loading and post-processing.

\begin{table}[h]
\centering
\caption{\textbf{Inference latency comparison on the ScanNet test set.} 
We report the wall-clock time averaged across all test scenes to process a full 90-frame sequence (\emph{Per-sequence}) and the amortized per-frame cost (\emph{Per-frame}).
All times measure model forward-pass latency only, excluding data loading and post-processing.
\emph{Offline}: all frames are available upfront and processed jointly.
\emph{Streaming}: frames arrive one at a time; each frame is processed incrementally using cached states.
\textsuperscript{\textdagger}~Offline per-frame latency is amortized: $\frac{\text{total time}}{\text{\# total frames}}$ in one sequence.}
\label{tab:streaming_timing}
\resizebox{0.9\linewidth}{!}{%
\begin{tabular}{l c cc cc}
\toprule
& & \multicolumn{2}{c}{\textbf{Offline}} 
& \multicolumn{2}{c}{\textbf{Streaming}} \\
\cmidrule(lr){3-4} \cmidrule(lr){5-6}
\textbf{Method} & \textbf{\#Params}
& Per-sequence (s) & Per-frame (s)\textsuperscript{\textdagger}
& Per-sequence (s) & Per-frame (s) \\
\midrule
VGGT~\cite{wang2025vggt} & 1.26B & 9.90 & 0.11 & --- & --- \\
CUT3R~\cite{wang2025continuous} & 0.80B & 5.22 & 0.06 & 6.03 & 0.07 \\
StreamVGGT~\cite{zhuo2026streaming} & 1.26B & --- & --- & 9.00 & 0.10 \\
\model (Ours) & 8.20B & \textbf{2.16} & \textbf{0.02} & \textbf{5.76} & \textbf{0.06} \\
\bottomrule
\end{tabular}%
}
\end{table}

As shown in \cref{tab:streaming_timing}, \model achieves the lowest latency in both offline and streaming settings despite having substantially more parameters. In offline mode, it reduces amortized per-frame latency to 0.02s, compared with 0.06s for CUT3R~\cite{wang2025continuous} and 0.11s for VGGT~\cite{wang2025vggt}. In streaming mode, it processes each frame in 0.06s, slightly faster than CUT3R and clearly faster than StreamVGGT.
We attribute this efficiency to three factors: (1) fewer visual tokens per frame from the SigLIP encoder~\cite{tschannen2025siglip}, (2) the lower-cost causal attention backbone, and (3) KV-cache reuse for incremental inference. More discussion of the inference setup and efficiency analysis is provided in \cref{app:latency_details}.

\section{Conclusion}

We introduce \model, a pose-grounded video understanding model that equips standard MLLMs with the capability to connect individual frames in a shared space. With a simple yet scalable architectural design and tailored training dynamics, \model improves spatial and general video QA, and achieves competitive streaming pose estimation performance against state-of-the-art methods. Our results position camera pose as an important missing signal for video MLLMs: it grounds frames in a globally consistent 3D space and encourages learning cross-frame correspondences. \model advances MLLMs toward real-world grounded video understanding.


\section*{Acknowledgments} 
We thank Oscar Michel, Baiqiao Yin, Jianyuan Wang, Anjali Gupta, Ellis Brown, Peter Tong, and Pinzhi Huang for reviewing this manuscript and providing constructive feedback. This work is supported by a grant from the Meta FAIR team. S.X. acknowledges support from the MSIT IITP grant (RS-2024-00457882) and the NSF award IIS-2443404.

\clearpage

\addcontentsline{toc}{section}{References}
\bibliography{arxiv}
\bibliographystyle{plain}

\clearpage
\appendix
\section*{Appendix}

\section{Implementation Details}
\label{app:implementation}

\subsection{Frame Sampling}
A key finding of \model is that camera pose estimation and video QA place fundamentally different demands on frame sampling. For VQA samples loaded from scene-frame sequences, we uniformly sample $N$ frame indices from the $L$ available frame entries, where $L\geq N$:
\begin{equation}
u_i = \left\lfloor (i-1) \cdot \frac{L-1}{N-1} \right\rfloor, \quad i = 1, \ldots, N.
\end{equation}
This ensures broad temporal coverage of the video content, which is essential for answering questions that may reference events at any point in the video. To mitigate memorization of fixed pose targets, we apply random jitter with $\Delta=\lfloor L\alpha\rfloor$ and $\alpha=0.005$. Setting $u_{N+1}=L$, we independently sample integer offsets $\delta_i \sim \operatorname{Unif}\{-\Delta,\ldots,\Delta\}$ and form clipped proposals
\begin{equation}
r_i=\operatorname{clip}\!\left(u_i+\delta_i,\,0,\,u_{i+1}-1\right), \quad i=1,\ldots,N.
\end{equation}
We then enforce temporal order recursively:
\begin{equation}
\tilde{u}_1=r_1, \qquad \tilde{u}_i=\max\!\left(r_i,\tilde{u}_{i-1}\right), \quad i=2,\ldots,N.
\end{equation}
Thus, the final indices satisfy $0\leq\tilde{u}_1\leq\cdots\leq\tilde{u}_N\leq L-1$.

\noindent\textbf{Dynamic temporal sampling.}
For dedicated pose-only samples during interleaved training, we adopt a two-mode sampling strategy following CUT3R~\cite{wang2025continuous}. With probability $p_\text{video}$, we sample in \emph{video mode}: a random starting frame is selected, and subsequent frames are drawn using either a fixed interval (with probability $p_{\text{fix}}$) or variable intervals uniformly sampled from $[I_{\text{min}}, I_{\text{max}}]$. With probability $1 - p_\text{video}$, we sample in \emph{collection mode}: frames are randomly drawn from the entire sequence.
The sampling parameters are dataset-specific to account for different frame rates and scene dynamics (see \cref{tab:dynamic_sampling_params}). The large interval ranges ensure diverse temporal baselines across training iterations, which is critical for robust pose estimation.

\begin{table}[h]
\centering
\small
\caption{\textbf{Dataset-specific sampling parameters for dynamic temporal sampling.}}
\label{tab:dynamic_sampling_params}
\begin{tabular}{lcccc}
\toprule
Dataset & $p_\text{video}$ & $p_\text{fix}$ & $I_{\text{min}}$ & $I_{\text{max}}$ \\
\midrule
ScanNet~\cite{dai2017scannet} & 0.6 & 0.6 & 30 & 100 \\
ScanNet++~\cite{yeshwanth2023scannet++} & 0.8 & 0.5 & 30 & 100 \\
ARKitScenes~\cite{dehghan2021arkitscenes} & 0.8 & 0.5 & 30 & 100 \\
\bottomrule
\end{tabular}
\end{table}

\subsection{MapAnything Training Setup}
\label{app:mapanything_setup}

For datasets from MapAnything~\cite{keetha2025mapanything}, we follow MapAnything's official sampling strategy to use covisibility-guided random walk sampling. Specifically, given a pre-computed pairwise covisibility matrix for each scene, we perform a random walk on the covisibility graph: starting from a random frame, at each step we move to a random unvisited neighbor whose normalized covisibility exceeds a threshold $\tau$ (dataset-specific, typically 0.15 $\sim$ 0.30). If no unvisited neighbor is available, we backtrack. This ensures that sampled frame sets form connected subgraphs with sufficient visual overlap for pose estimation, while maintaining diversity. If the desired number of frames cannot be reached, we attempt up to four restarts, excluding previously visited components.

For the camera pose estimation experiments presented in \cref{sec:pose_exp}, we utilize a mixture of pose-annotated data from \vsidata (\ie, ScanNet~\cite{dai2017scannet}, ScanNet++~\cite{yeshwanth2023scannet++}, and ARKitScenes~\cite{dehghan2021arkitscenes}) alongside datasets from MapAnything~\cite{keetha2025mapanything}. The latter includes metric-scale datasets (ParallelDomain4D~\cite{van2024generative}, TartanAir-v2~\cite{wang2020tartanair,zhang2025ufm}, MVS-Synth~\cite{huang2018deepmvs}, Spring~\cite{mehl2023spring}, SailVOS3D~\cite{hu2021sail}, ETH3D~\cite{schops2017multi}, Dynamic Replica~\cite{karaev2023dynamicstereo}, MPSD~\cite{antequera2020mapillary}, and UnrealStereo4K~\cite{tosi2021smd}) as well as non-metric-scale datasets (MegaDepth~\cite{li2018megadepth}, DL3DV~\cite{ling2024dl3dv}, and BlendedMVS~\cite{yao2020blendedmvs}). Since VQA annotations are available for the ScanNet, ScanNet++, and ARKitScenes subsets of \vsidata, the original VQA samples retain the NTP loss and use uniform sampling. Their interleaved pose-only counterparts use CUT3R-style dynamic temporal sampling~\cite{wang2025continuous}, while all MapAnything datasets use covisibility-based sampling~\cite{keetha2025mapanything}.

\subsection{Pseudo-Pose Annotation Pipeline}
\label{app:pseudo_pose_pipeline}

To extend pose supervision beyond \vsidata{}'s GT-pose subset, we annotate \cambrians{}-3M~\cite{yang2026towards}, the open-source video instruction-tuning corpus used in the Stage~3 general-video training of \cambrians{}. \cambrians{}-3M aggregates LLaVA-Video-178K~\cite{zhang2024video}, LLaVA-Hound / ShareGPTVideo~\cite{zhang2024direct}, and an additional NYU-curated portion, drawing from $\sim$30 underlying open-domain video sources including Kinetics-400/600/700~\cite{kay2017kinetics,carreira2018short,carreira2019short}, NTU-RGBD~\cite{shahroudy2016ntu}, ActivityNet~\cite{caba2015activitynet}, Ego4D~\cite{grauman2022ego4d}, EpicKitchens~\cite{damen2018scaling}, LSMDC~\cite{rohrbach2015dataset}, Something-Something-V2~\cite{goyal2017something}, WebVid~\cite{bain2021frozen}, NextQA~\cite{xiao2021next}, Vript~\cite{yang2024vript}, GUI-World~\cite{chen2024gui}, and others. Each retained clip is annotated with per-frame camera extrinsics (translation $\mathbf{t}$ and rotation quaternion $\mathbf{q}$) and intrinsics (FoV), matching the 9-D pose encoding consumed by $\mathcal{L}_{\text{pose}}$ in \cref{sec:train_objective}. The pipeline runs two filtering stages followed by VIPE pose annotation and post-filtering.

\paragraph{Stage 1: Scene-cut detection.}
Pose estimation assumes a single continuous camera trajectory, so clips containing scene cuts are excluded up front. We run PySceneDetect's \texttt{ContentDetector} (HSV-histogram content threshold $45.0$) augmented with a frame-level histogram-Bhattacharyya check (threshold $0.65$), retaining only single-scene clips of at least 3 seconds.

\paragraph{Stage 2: Pose-aware VLM filtering.}
Surviving clips are screened by Qwen3-VL~\cite{Qwen3-VL} using the prompt shown in the box below. The prompt asks the VLM nine yes/no questions, including seven hard rejection criteria---synthetic or animated content, large text overlays, screen recordings, severe blur or focus loss, heavy compression, extreme exposure, and shot-through-glass reflections---and two metadata-only flags for downstream analysis: dynamic-scene-only and low-parallax. A clip is discarded if any hard rejection criterion is triggered.

\begin{tcolorbox}[
    enhanced,
    colback=gray!4,
    colframe=gray!55,
    title={\small\textbf{Stage~2 VLM filtering prompt}},
    fonttitle=\normalsize,
    fontupper=\footnotesize,
    boxrule=0.4pt,
    arc=2pt,
    left=5pt, right=5pt, top=4pt, bottom=4pt,
    boxsep=2pt,
    breakable,
]
Analyze this video frame for suitability in a camera pose estimation dataset. Be conservative: only flag if you are HIGHLY confident ($>\!90\%$).

\smallskip
Answer the following questions:
\begin{enumerate}[leftmargin=1.8em, topsep=2pt, itemsep=0pt, parsep=0pt]
\item \texttt{SYNTHETIC}: CGI, animation, video game, or computer-generated? (not real camera footage)
\item \texttt{TEXT\_OVERLAY}: subtitles, captions, large watermarks, or title cards?
\item \texttt{SCREEN\_RECORDING}: a recording of a screen / monitor / TV?
\item \texttt{BLUR}: severely blurry or out-of-focus such that edges / landmarks are not usable for pose estimation? (Do not flag mild motion blur.)
\item \texttt{COMPRESSION}: heavily pixelated / blocky / very low resolution such that details are lost?
\item \texttt{LIGHTING}: extremely dark or severely overexposed such that scene structure is not visible?
\item \texttt{REFLECTION/THROUGH-GLASS}: mostly shot through glass (windshield / window) with strong reflections / glare obscuring the scene?
\item \texttt{DYNAMIC\_SCENE} (metadata only): mostly moving objects (crowd / traffic) with few static landmarks?
\item \texttt{LOW\_PARALLAX} (metadata only): near-zero parallax (rotation-only or tiny motion) making geometry-based pose estimation unreliable?
\end{enumerate}
\smallskip
Return a JSON object with boolean fields for each question and set \texttt{suitable\_for\_pose}{=}\texttt{FALSE} if any of (1)--(7) is true.
\end{tcolorbox}

\paragraph{Stage 3: ViPE pose annotation.}
Filtered clips are processed by VIPE~\cite{huang2025vipe}, a recent feed-forward streaming video pose engine, which produces per-frame extrinsics and intrinsics. We retain only the pose track and discard auxiliary outputs (dense depth, point clouds) since downstream training consumes only $[\mathbf{t}_i, \mathbf{q}_i, f_i^h, f_i^w]$. Clips that failed on the VIPE pipeline (e.g., numerical instability on very short or content-poor sequences) are discarded.

\paragraph{Incorporating into training.}
Pseudo-pose samples are routed through the same training dynamics as GT-pose samples: pose-only augmented samples carry only $\mathcal{L}_{\text{pose}}$, and joint VQA+pose samples carry both losses. Since pseudo-pose quality varies across video sources, we apply source-level filtering during training and retain only sources that produce stable VIPE trajectories under our annotation pipeline. We do not introduce any architecture change or special loss weighting for pseudo poses. The improvement reported in \cref{tab:general_vqa} therefore reflects the benefit of additional pose-supervised data.

\subsection{Additional Latency Details}
\label{app:latency_details}

Although \model contains significantly more parameters than previous specialist 3D reconstruction models, it achieves strong inference efficiency for camera pose estimation due to its compact visual representation, causal transformer backbone, and the highly optimized inference infrastructure of the underlying LLM. Here we provide additional details complementing the main-text results in \cref{tab:streaming_timing}. All latency measurements are conducted on the ScanNet~\cite{dai2017scannet} test set using a single NVIDIA L40S GPU, excluding data loading and post-processing.

\paragraph{Offline mode.}
When all frames are available upfront, \model processes them in a single LLM forward pass with full KV-cache prefill, achieving an amortized per-frame cost of only 0.02s. This is approximately $3\times$ faster than CUT3R~\cite{wang2025continuous} at 0.06s per frame and $5.5\times$ faster than VGGT~\cite{wang2025vggt} at 0.11s per frame. In this setting, the model benefits from jointly processing the full sequence while still using a compact token budget per frame.

\paragraph{Streaming mode.}
When frames arrive sequentially, \model processes each new frame at 0.06s per frame, slightly faster than CUT3R~\cite{wang2025continuous} at 0.07s and clearly faster than StreamVGGT~\cite{zhuo2026streaming} at 0.10s. In this setting, each new frame's visual and pose tokens attend to the cached KV states of previously processed frames, avoiding recomputation over the full history. CUT3R achieves comparable speed through a recurrent state mechanism that carries a fixed-size memory forward, while VGGT does not support streaming because its bidirectional attention requires access to all frames simultaneously.

\paragraph{Method-specific remarks.}
VGGT~\cite{wang2025vggt} uses bidirectional attention over all frames and therefore only supports offline inference. StreamVGGT~\cite{zhuo2026streaming} is streaming-native and does not support offline joint processing. CUT3R~\cite{wang2025continuous} supports both modes: in offline mode, it batch-encodes all frames through its ViT encoder and then sequentially steps through its recurrent decoder; in streaming mode, it processes each incoming frame with a single encode-decode update.

\paragraph{Efficiency analysis.}
We attribute the practical efficiency of \model to three factors. First, \emph{compact visual representation}: \model uses substantially fewer visual tokens per frame than the DINOv2-based~\cite{oquab2023dinov2} encoders adopted by VGGT~\cite{wang2025vggt} and CUT3R~\cite{wang2025continuous}, directly reducing attention cost. Second, \emph{causal transformer architecture}: the causal attention mask yields lower computation than bidirectional attention over the same sequence length. Third, \emph{KV-cache reuse}: standard causal LLM inference avoids recomputing attention over previous frames, which is especially beneficial in the streaming setting.

Overall, these results show that \model can combine strong spatial reasoning, competitive streaming pose estimation, and favorable inference speed despite its substantially larger parameter count.

\section{Additional Ablations}
\label{app:additional_ablations}

\subsection{Evaluation on ReVSI}

\label{app:revsi}
We further evaluate \model and its
no-pose counterpart on ReVSI~\cite{zhang2026revsi}  and compare against representative proprietary,
general-purpose open-source, and spatial-specialist open-source models. ReVSI rebuilds visual spatial intelligence evaluation with expert annotations
and frame-adaptive ground-truth answers, addressing annotation noise in the original VSI-Bench. Notice that \model evaluation happens only on ReVSI-All, as we believe the benchmark should measure whether an MLLM can correctly answer questions about the video, independent of how the evaluation inputs are concretely configured.

As shown in \cref{tab:revsi}, \model achieves the best performance among open-source models of comparable size. Pose supervision also continues to provide substantial gains over the no-pose baseline. However, \model exhibits smaller gains on ReVSI than on VSI-Bench. We attribute this to two factors. First, there is a frame-sampling mismatch. ReVSI provides frame-adaptive ground-truth answers for specific frame budgets, whereas our strongest VSI-Bench setting uses 128 frames, for which ReVSI does not provide directly corresponding frame-adaptive ground-truth. As a result, the 128-frame comparison is less well aligned with the ReVSI evaluation protocol.
Second, \model is trained on in-distribution VSI-Bench data, \ie, \vsidata, so its prediction distribution is naturally better matched to VSI-Bench. Although ReVSI evaluates the same videos, it uses a different annotation protocol. Models fine-tuned on VSI-Bench in-distribution data may therefore experience a distribution shift when evaluated on ReVSI. This behavior is expected and intuitive.

\begin{table*}[!t]
\centering
\caption{\textbf{ReVSI Results Comparison}. We report ReVSI average and per-subcategory results. ReVSI uses frame-adaptive ground-truth answers under each model's inference frame setting. \model here is trained with \vsidata and a 590K subset of Cambrian-S-3M.}
\label{tab:revsi}
\resizebox{1\textwidth}{!}{%
\begin{tabular}{ll|c|cccc|ccc}
\toprule
\multirow{2}{*}{Model} & \multirow{2}{*}{Frames} & & \multicolumn{4}{c|}{Numerical Answer} & \multicolumn{3}{c}{Multiple-Choice Answer} \\
\cmidrule(lr){4-7} \cmidrule(lr){8-10}
& & Avg. & Obj. Count & Abs. Dist. & Obj. Size & Room Size & Rel. Dist. & Rel. Dir. & Route Plan \\
\midrule
\rowcolor{gray!10!blue!5}\textit{Proprietary Models} & & & & & & & & & \\
GPT-5.2 & 64 & 50.9 & 56.2 & 41.5 & 73.9 & \textbf{63.0} & 48.4 & 34.9 & 38.2 \\
Gemini 3 Flash & 1 FPS & 57.6 & \textbf{65.7} & 53.1 & 77.6 & 52.8 & 64.6 & 47.9 & 41.8 \\
Gemini 3 Pro & 1 FPS & \textbf{60.9} & 60.1 & 54.7 & \textbf{79.3} & 51.9 & \textbf{68.1} & \textbf{56.0} & \textbf{56.4} \\
\midrule
\rowcolor{gray!10!blue!5}\textit{Open-source Models} & & & & & & & & & \\
Qwen3-VL-8B-Instruct~\cite{Qwen3-VL} & 64 & 49.1 & 40.4 & 52.3 & \textbf{69.0} & 45.1 & \textbf{57.1} & 39.5 & 40.5 \\
Qwen2.5-VL-7B-Instruct~\cite{bai2025qwen2} & 4 FPS & 32.6 & 36.9 & 15.0 & 49.7 & 29.0 & 31.5 & 29.5 & 36.7 \\
InternVL-3.5-8B~\cite{wang2025internvl3} & 64 & 47.9 & 43.3 & 54.6 & 64.2 & \textbf{47.6} & 45.0 & 36.3 & 44.4 \\
LLaVA-Video-7B-Qwen2~\cite{zhang2024video} & 64 & 30.3 & 31.3 & 1.4 & 52.5 & 16.7 & 38.3 & 33.3 & 38.4 \\
\cambrians-7B~\cite{yang2026towards} & 128 & 49.1 & \textbf{48.4} & 60.5 & 65.5 & 46.7 & 37.1 & 48.5 & 37.0 \\
VST-7B-SFT~\cite{yang2025visual} & 4 FPS & 46.4 & 35.4 & 52.6 & 67.9 & 47.2 & 49.2 & 36.9 & 35.4 \\
\model\ (w/o Pose) & 128 & 50.1 & 42.3 & 64.6 & 64.5 & 47.4 & 38.1 & 48.3 & 45.3 \\
\rowcolor{green!5}\textbf{\model} & 128 & \textbf{52.0} & 41.4 & \textbf{68.3} & 66.7 & 45.9 & 40.2 & \textbf{48.7} & \textbf{53.1} \\
\bottomrule
\end{tabular}
}
\end{table*}

\subsection{Depth-Baseline Fairness}
\label{app:depth_fairness}
The loss ablation in \cref{tab:loss_ablation} compares pose and depth supervision under the same training recipe: interleaved training with CUT3R-style augmentation~\cite{wang2025continuous}. To verify that the pose-vs-depth gap is not caused by an under-tuned depth baseline, we further sweep the key recipe choices for depth supervision: whether to use interleaved training, whether to apply augmentation used in CUT3R~\cite{wang2025continuous}, and the depth-loss weighting factor $\lambda_d$.

\begin{table}[h]
\centering
\scriptsize
\caption{\textbf{Depth-baseline fairness ablation on VSI-Bench.}
All experiments use \vsidata with 32 frames and 196 visual tokens per frame.
(a) Fixing $\lambda_d{=}1.0$, we ablate interleaved training and data augmentation.
(b) With interleaved training and data augmentation enabled, we sweep the depth-loss weight $\lambda_d$ under both depth-only and depth+pose supervision setups.
}
\label{tab:depth_appendix}
\begin{minipage}[t]{0.49\linewidth}
\centering
\textit{(a) Recipe sweep, depth-only}\\[-1pt]
\begin{tabular}{lcc}
\toprule
& Aug ON & Aug OFF \\
\midrule
Interleaved ON  & 69.4 & 69.6 \\
Interleaved OFF & 69.0 & 70.1 \\
\bottomrule
\end{tabular}
\end{minipage}
\begin{minipage}[t]{0.49\linewidth}
\centering
\textit{(b) $\lambda_d$ sweep}\\[-1pt]
\begin{tabular}{lccc}
\toprule
$\lambda_d$ & 0.2 & 1.0 & 5.0 \\
\midrule
Depth       & 70.6 & 69.4 & 67.0 \\
Depth+Pose  & 71.4 & 71.7 & 69.9 \\
\bottomrule
\end{tabular}
\end{minipage}
\end{table}

As shown in \cref{tab:depth_appendix}(a), we ablate two recipe choices for the depth-only experiments: interleaved training and pose-style augmentation, while fixing the depth-loss weight to $\lambda_d{=}1.0$. The results have only subtle changes across these settings, suggesting that the depth-only underperformance is not caused by these two design choices. \cref{tab:depth_appendix}(b) then tests whether the depth objective is under- or over-weighted, sweeping $\lambda_d$ for both depth-only and depth+pose supervision under the default recipe. Reducing the depth weight improves the depth-only baseline. However, all depth-only variants remain below the pose-only result of 72.0 reported in \cref{tab:loss_ablation}. Adding depth supervision on top of pose also fails to close the gap: the best depth+pose variant is still slightly below the pose-only supervision baseline.

These results suggest that the advantage of pose supervision is not merely due to insufficient depth tuning, but reflects a better alignment between the camera pose estimation and the video understanding. 
We attribute the suboptimal performance of depth supervision to two factors. First, depth is a dense, per-pixel prediction target, which introduces optimization challenges for a video LLM that represents each frame with only 64 $\sim$ 196 visual tokens. Second, following VGGT, we use per-frame depth supervision, which primarily captures local scene geometry within each individual frame. In contrast, camera pose directly specifies how different views relate to one another in a shared coordinate frame, which therefore provides a compact global signal for cross-frame video reasoning.

\subsection{Pose-Data Scalability}
\label{app:pose_data_efficiency}
How does \model scale with the amount of pose-annotated data? Within \vsidata, approximately 49\% of the training pairs carry pose annotations. We sweep the pose-annotated fraction from 0\% to 49\% in five steps (corresponding to 0\%, 25\%, 50\%, 75\%, and 100\% of the available pose pairs) and measure both VSI-Bench accuracy and ScanNet ATE under two frame-count settings.

\begin{figure}[h]
\centering
\includegraphics[width=0.78\textwidth]{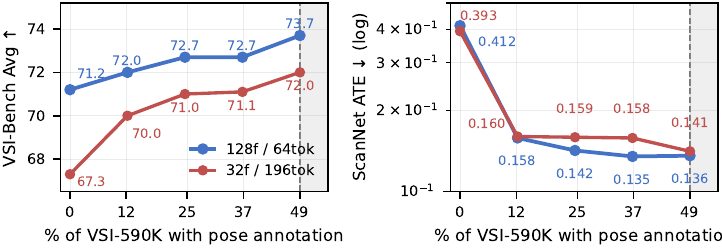}
\caption{\textbf{Pose-data scalability.} VSI-Bench Avg (left) and ScanNet ATE (right) as the pose-annotated fraction of \vsidata varies from 0\% (pure VQA) to 49\% (the cap, since the remaining \vsidata{} samples are image-only). Curves are shown for two frame settings: 128f / 64tok and 32f / 196tok. Both VQA and pose accuracy improve as more pose-annotated data is added.}
\label{fig:pose_data_efficiency}
\end{figure}

\cref{fig:pose_data_efficiency} shows that both VSI-Bench accuracy and ScanNet ATE improve as the pose-annotated fraction grows, with the largest jump occurring between 0\% and the first non-zero point and gains continuing through 49\%. The 128f / 64tok curve dominates the 32f / 196tok curve at every fraction, consistent with the frame-count scaling reported in \cref{tab:vsi_frame}. These trends suggest \model{} can take advantage of additional pose data well beyond what \vsidata currently provides, motivating both the pseudo-pose training in \cref{tab:general_vqa} and the MapAnything-scale training in \cref{sec:pose_exp}.

\section{Visualizations}
\label{app:visualization}

\subsection{Additional Camera Pose Trajectory Visualizations}
\label{app:scannet_val_visualization}

\begin{figure}[h]
\centering
\includegraphics[width=\textwidth]{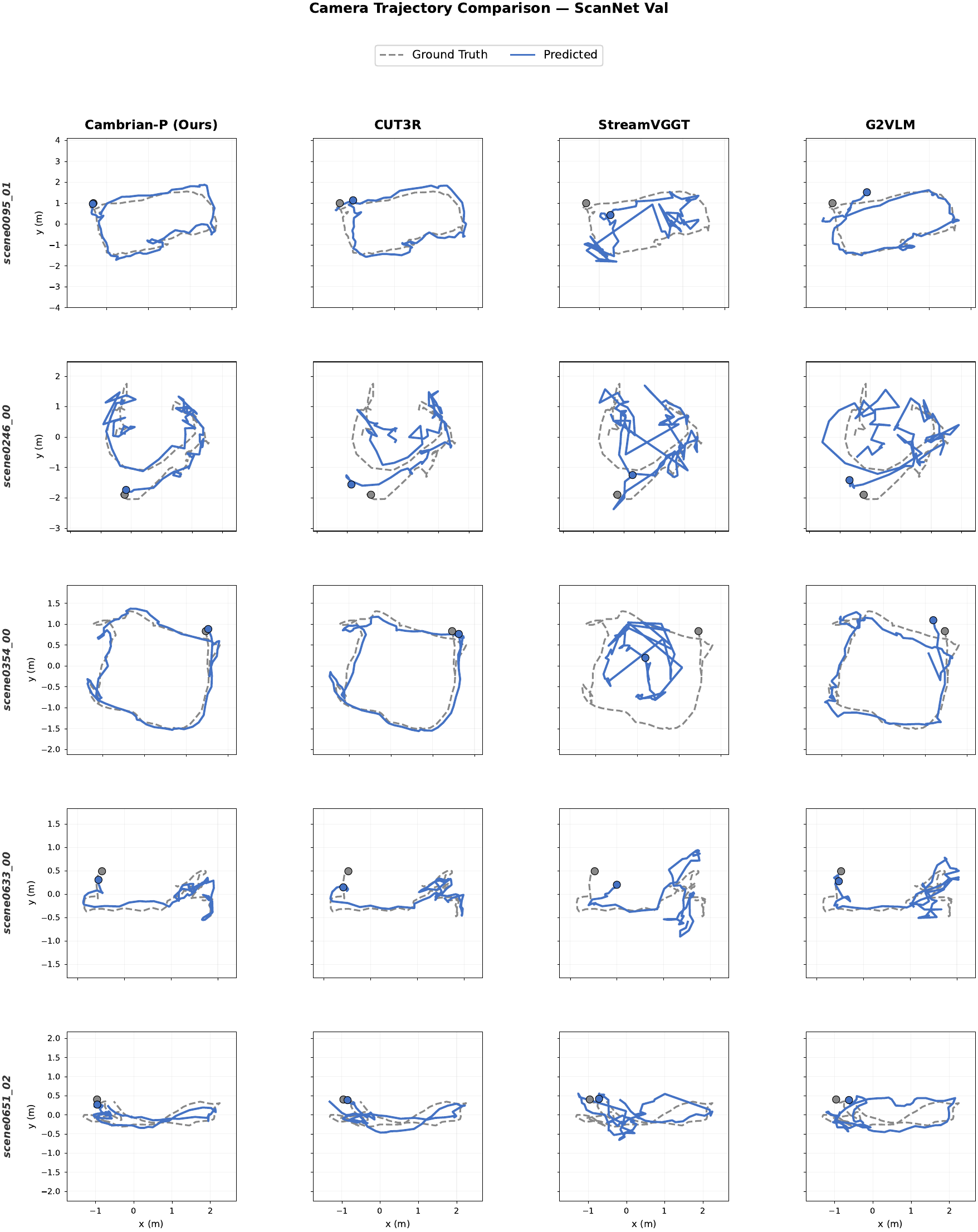}
\vspace{-0.5cm}
\caption{\textbf{Camera pose trajectory visualization on ScanNet validation scenes (1/2).} Ground-truth trajectories are shown in gray dashed lines and predicted trajectories in blue solid lines. Each column corresponds to a different method. These scenes overlap with VSI-Bench~\cite{yang2024think} evaluation scenes.}
\label{fig:scannet_traj_val1}
\end{figure}

\begin{figure}[h]
\centering
\includegraphics[width=\textwidth]{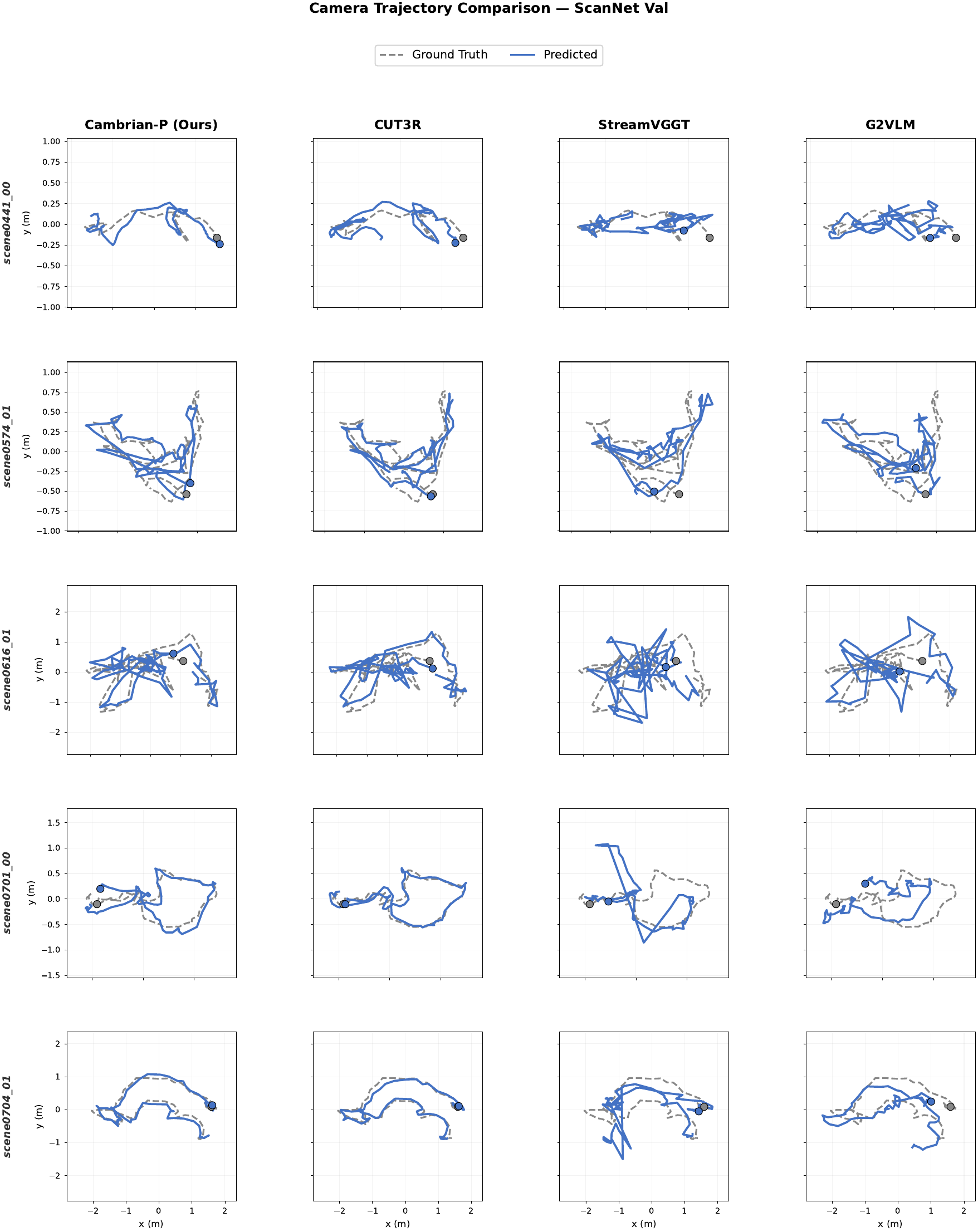}
\vspace{-0.5cm}
\caption{\textbf{Camera pose trajectory visualization on ScanNet validation scenes (2/2).} Continued from \cref{fig:scannet_traj_val1}.}
\label{fig:scannet_traj_val2}
\end{figure}

We provide additional qualitative camera pose trajectory comparisons on ScanNet validation scenes in \cref{fig:scannet_traj_val1,fig:scannet_traj_val2}. These scenes overlap with the subset used in VSI-Bench~\cite{yang2024think}. For each scene, we plot the ground-truth trajectory (gray dashed) and the aligned predicted trajectories (blue solid) from \model, CUT3R~\cite{wang2025continuous}, StreamVGGT~\cite{zhuo2026streaming}, and G$^2$VLM~\cite{hu2025g}. Consistent with the main-text results on the ScanNet test set, \model recovers trajectory shapes that closely match the ground truth across diverse scenes.

\subsection{OOD Pose Trajectories on EgoSchema}
\label{app:ood_pose_egoschema}

To understand how \model{}'s pose head generalizes outside its training distribution, we visualize trajectories predicted by \model trained on VSI-590K on EgoSchema~\cite{mangalam2023egoschema} clips. EgoSchema contains long-form egocentric videos that are disjoint from the indoor / synthetic scenes used in \vsidata and MapAnything~\cite{keetha2025mapanything} training, and has no metric pose ground truth. We use VIPE~\cite{huang2025vipe} pseudo-GT trajectories as the reference for comparison.

\begin{figure}[h]
\centering
\includegraphics[width=0.92\textwidth]{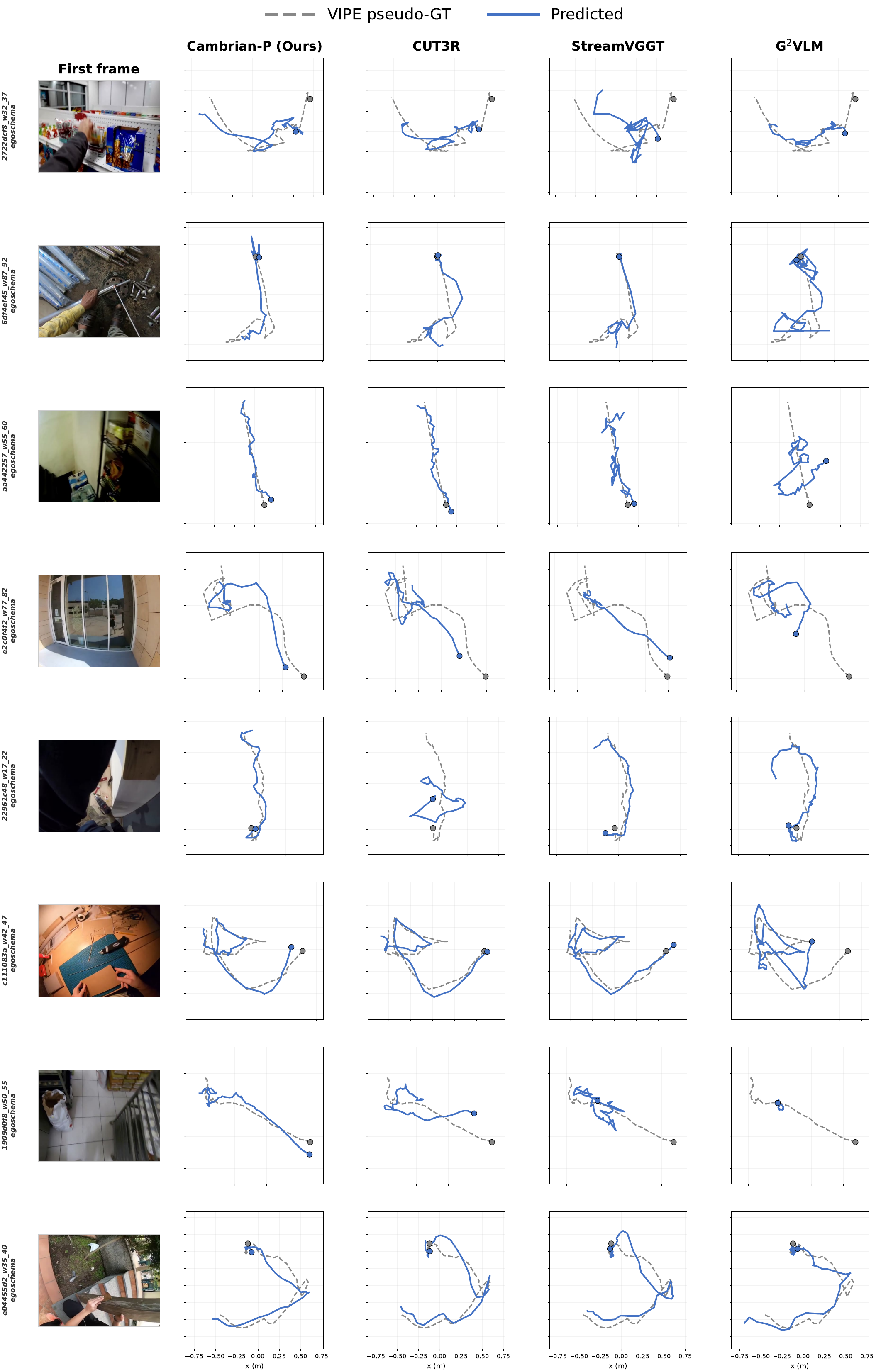}
\caption{\textbf{OOD pose trajectories on EgoSchema.} Pseudo-GT trajectories annotated by VIPE~\cite{huang2025vipe} are shown in gray dashed lines and predicted trajectories in blue solid lines. }
\label{fig:ood_pose_egoschema}
\end{figure}

\cref{fig:ood_pose_egoschema} shows eight scenes. Across all scenes, \model{}'s predicted trajectory better tracks the overall shape and scale of the VIPE pseudo-GT than the specialist baselines, despite \model{} being trained only on \vsidata and MapAnything data. This complements the in-distribution ScanNet results in \cref{fig:scannet_traj_test} and indicates that pose supervision within an MLLM yields a generalization-ready geometric prior rather than a domain-specific pose regressor.

\subsection{VQA Qualitative Examples}
\label{app:vqa_viz}

\begin{figure}[h]
\centering
\vspace{-0.5cm}
\includegraphics[width=0.9\textwidth,page=1]{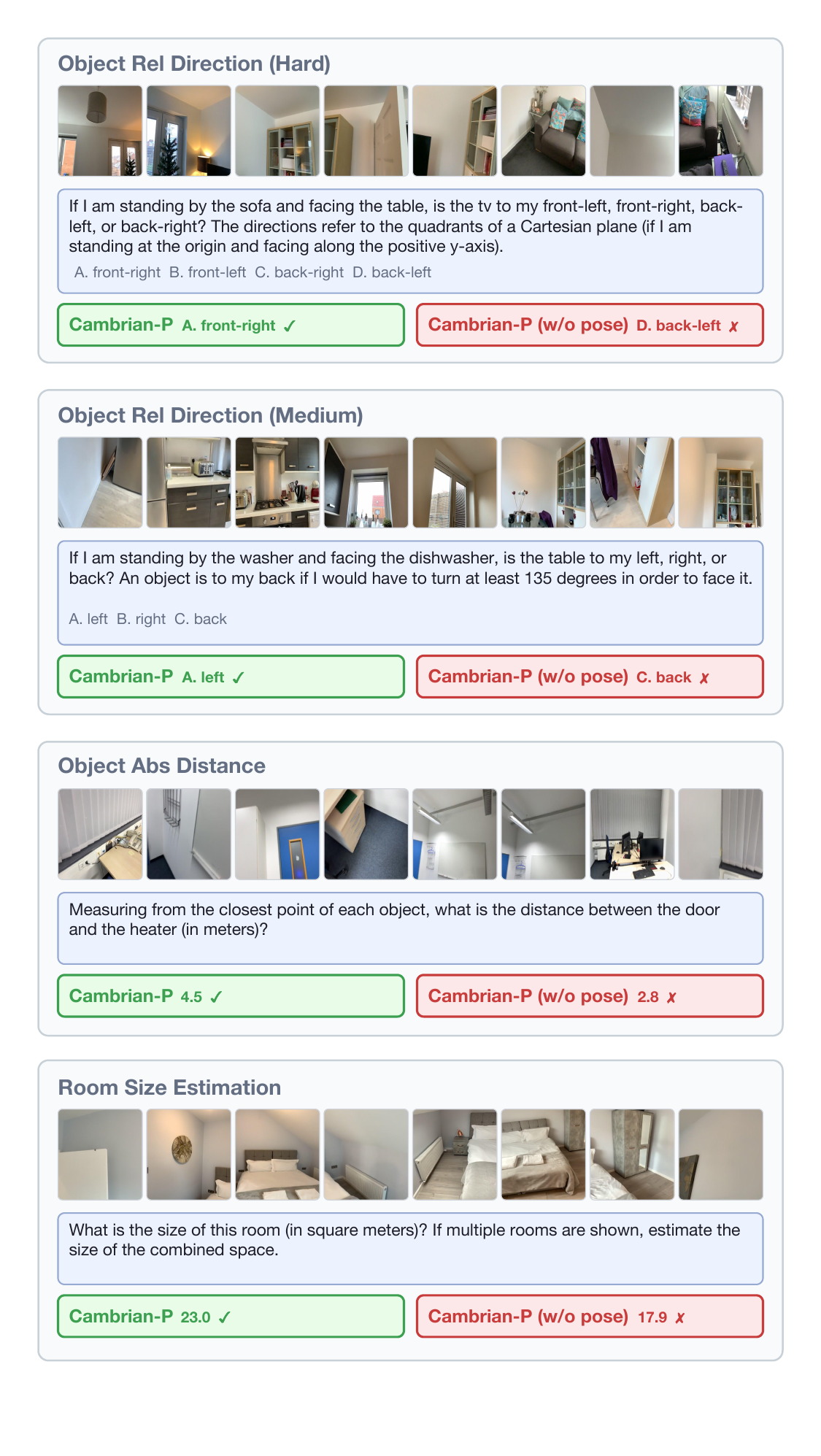}
\vspace{-1.8cm}
\caption{\textbf{Qualitative VQA comparison (1/2).} We compare \model and \model (w/o pose) on VSI-Bench spatial reasoning questions.}
\label{fig:vqa_qualitative_1}
\end{figure}

\begin{figure}[h]
\vspace{-0.5cm}
\centering
\includegraphics[width=0.9\textwidth,page=2]{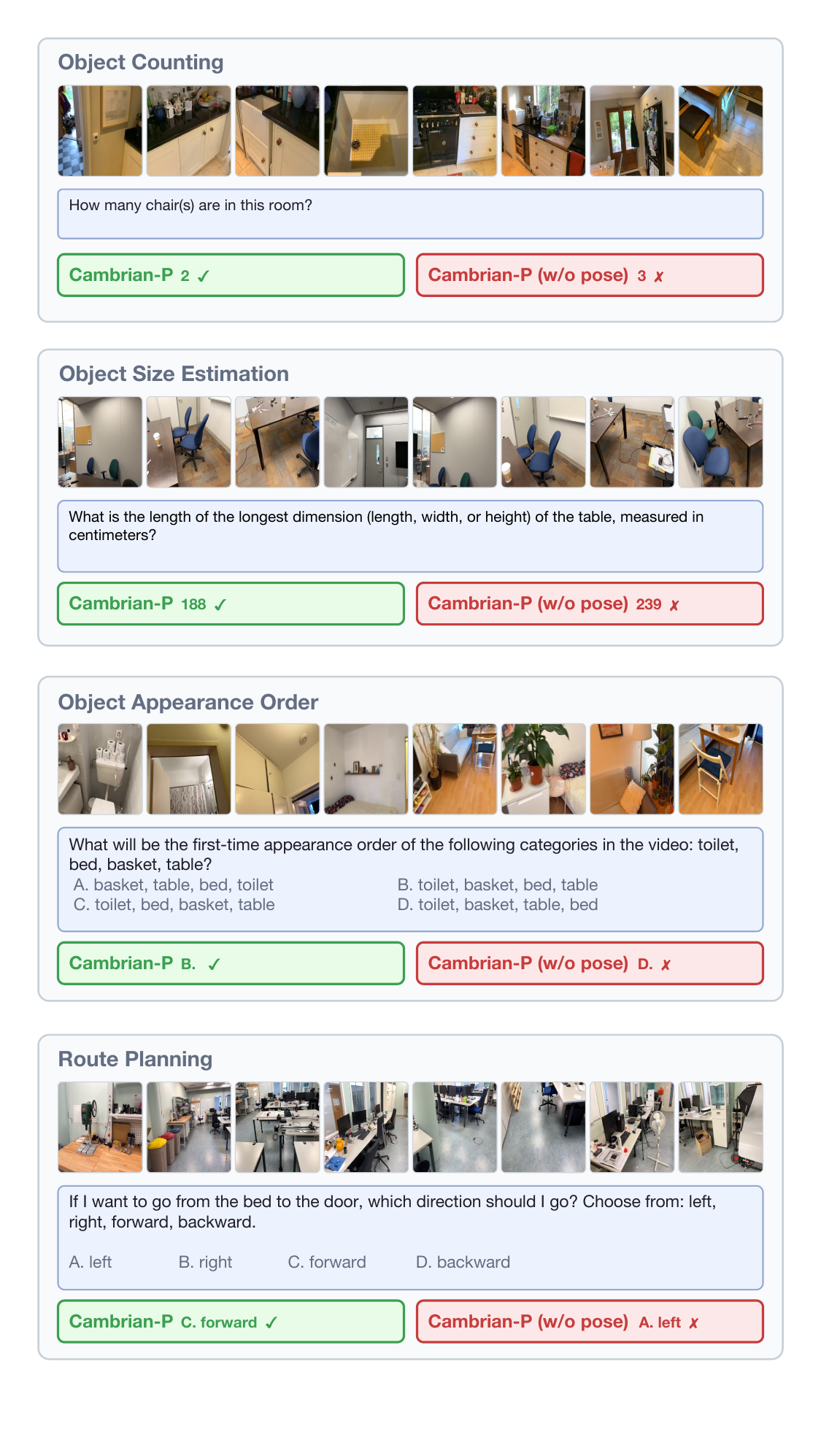}
\vspace{-1.8cm}
\caption{\textbf{Qualitative VQA comparison (2/2).} Continued from \cref{fig:vqa_qualitative_1}.}
\label{fig:vqa_qualitative_2}
\end{figure}

We show qualitative examples comparing \model with its no-pose counterpart (\model w/o pose) on spatial reasoning questions from VSI-Bench~\cite{yang2024think} in \cref{fig:vqa_qualitative_1} and \cref{fig:vqa_qualitative_2}. Each example shows a sequence of video frames, the question, and the answers from both models. The examples span all eight VSI-Bench subtasks: object relative direction (hard and medium difficulty), absolute distance estimation, room size estimation, object counting, object size estimation, appearance order, and route planning.

\end{document}